\documentclass[preprint,12pt,authoryear]{elsarticle}

\usepackage{lineno,amsmath, amsthm, amscd,bm, amsfonts,longtable, amssymb, graphicx,multirow,geometry,rotating, url, float, makecell}
\usepackage[colorlinks]{hyperref}
\usepackage{blindtext}

\usepackage{booktabs}
\usepackage{subcaption,graphicx,placeins}
\usepackage[usenames, dvipsnames]{color}
\usepackage[british]{babel}
\usepackage{lipsum,array, afterpage,pdflscape,booktabs}
\modulolinenumbers[5]
\makeatletter \oddsidemargin.9375in \evensidemargin \oddsidemargin
\theoremstyle{definition}

\theoremstyle{remark}

\numberwithin{equation}{section}

\journal{A}
\allowdisplaybreaks
\begin{document}

\begin{frontmatter}
\title{Physics-Informed Neural Networks with Unknown Partial Differential Equations: an Application in Multivariate Time Series}
\author{Seyedeh Azadeh Fallah Mortezanejad$^1$}
\author{Ruochen Wang$^{*2}$}
\author{Ali Mohammad-Djafari$^{3,4}$}
\address{
$^{1,2}$ School of Automotive and Traffic Engineering, Jiangsu University, Zhenjiang, Jiangsu, China.\\
$^{3}$ International Science Consulting and Training (ISCT), 91440 Bures sur Yvette, France.\\
$^{4}$ Shanfeng Company, Shaoxing 312352, China.\\
$^*$ Contact: wrc@ujs.edu.cn. }

\begin{abstract}
A significant advancement in Neural Network (NN) research is the integration of domain-specific knowledge through
custom loss functions. This approach addresses a crucial challenge: how can models utilize
physics or mathematical principles to enhance predictions when dealing with sparse, noisy, or
incomplete data? Physics-Informed Neural Networks (PINNs) put this idea into practice by incorporating physical equations, such as
Partial Differential Equations (PDEs), as soft constraints. This guidance helps the networks find solutions that align with
established laws. Recently, researchers have expanded this framework to include Bayesian NNs (BNNs), which allow
for uncertainty quantification while still adhering to physical principles. But what happens when
the governing equations of a system are not known? In this work, we introduce methods to
automatically extract PDEs from historical data. We then integrate these learned equations into
three different modeling approaches: PINNs, Bayesian-PINNs (B-PINNs), and Bayesian Linear Regression (BLR). To assess these frameworks, we
evaluate them on a real-world Multivariate Time Series (MTS) dataset. We compare their effectiveness in forecasting future
states under different scenarios: with and without PDE constraints and accuracy considerations.
This research aims to bridge the gap between data-driven discovery and physics-guided learning,
providing valuable insights for practical applications.
\end{abstract}
\begin{keyword}
Physics-Informed Neural Network (PINN) \sep Bayesian Computation \sep Partial Differential
Equations (PDEs) \sep Multivariate Time Series (MTS).
\end{keyword}

\end{frontmatter}

\section{Introduction}

NN algorithms are widely used across various fields for a multitude of tasks,
often outperforming other methods. Consequently, a diverse range of
algorithms has been developed. For instance, there are algorithms designed
for spatial relationships in images, sequential data like MTS, and
tree-based models demonstrating impressive performance. Most of these
algorithms require rich datasets to effectively learn patterns and features
for tasks such as regression, classification, or clustering. PINNs, first
introduced by \cite{raissi2019physics}, improve upon traditional methods by
incorporating known physics—specifically PDEs and boundary conditions—into
the loss functions as a regularization technique, resulting in more accurate
outcomes. One of the key benefits of merging data-driven approaches with
fundamental physical principles is their effectiveness in handling small
sample sizes and noisy data, which makes them faster and more user-friendly.
Their versatility allows them to be applied across various fields, including
fluid dynamics, heat transfer, and structural analysis, where conventional
numerical methods often face challenges.

\cite{bararnia2022application} explored the application of PINNs to address
viscous and thermal boundary layer problems. They analyzed three benchmark
scenarios: Blasius-Pohlhausen, Falkner-Skan, and natural convection,
specifically investigating the impact of equation nonlinearity and unbounded
boundary conditions on the network architecture. Their findings revealed that
the Prandtl number plays a crucial role in determining the optimal number of
neurons and layers required for accurate predictions using TensorFlow.
Additionally, the trained models effectively predicted boundary layer
thicknesses for previously unseen data. \cite{hu2024physics} presented an
interpretable surrogate solver for computational solid mechanics based on
PINNs. They addressed the challenges of incorporating imperfect and sparse
data into current algorithms, as well as the complexities of high-dimensional
solid mechanics problems. The paper also explores the capabilities,
limitations, and emerging opportunities for PINNs in this domain.
\cite{wang2022applications} explored one application of PINNs in optical
fiber communication, proposing five potential solutions for modeling in the
time, frequency, and spatial domains. They modeled both forward and backward
optical waveform propagation by solving the nonlinear Schrödinger equation.
Although conventional numerical methods and PINNs produced similar accuracy,
PINNs demonstrated lower computational complexity and reduced time
requirements for the same optical fiber problems.

B-PINNs are an extension of PINNs that incorporate Bayesian inference to quantify uncertainty in
predictions. They leverage the strengths of NNs, physics-based modeling, and Bayesian methods to
effectively solve PDEs and other scientific problems. One of the key advantages of B-PINNs is their
ability to provide measures of uncertainty, which are crucial for risk assessment and informed
decision-making. \cite{liang2024higher} introduced a higher-order spatiotemporal
physics-incorporated graph NN to tackle missing values in MTS. By integrating a dynamic Laplacian
matrix and employing an inhomogeneous PDE, their approach effectively captured complex
spatiotemporal relationships and enhanced explainability through normalizing flows. Experimental
results demonstrated that this method outperformed traditional data-driven models, offering
improved dynamic analysis and insights into the impact of missing data.

Since the introduction of PINNs, they have been applied to a wide range of
problems across various fields. One of the notable advantages of PINNs is their speed and
effectiveness when working with small, noisy datasets.
However, a key question arises: how can we apply PINNs when we do not have PDEs readily available for our specific problems?
While many physical laws governing phenomena in physics, chemistry, biology, engineering, and environmental science are well-established, there are scenarios where no such laws or governing equations (e.g., PDEs) are known. In these cases, applying PINNs becomes challenging because PINNs rely on known physical laws to guide the learning process and achieve precise results with smaller sample sizes, fewer layers, fewer neurons, and fewer training epochs.

For example, consider sociological data, where there are no mathematical equations to describe the percentages of different behaviors or trends in human populations. Similarly, in economics, finance, psychology, or social media analytics, the relationships between variables are often complex and not easily captured by predefined PDEs. In such scenarios, one might aim to use PINNs to inform the network about future predictions in regions where no data is available, enabling the network to make precise forecasts even in data-sparse regions.

%
However, if no PDEs are known or applicable to the data, using PINNs directly is impossible. This highlights the importance of investigating suitable methods for extracting PDEs from raw data. By identifying governing equations or relationships within the data, we can bridge the gap between data-driven methods and physics-informed approaches, enabling the application of PINNs even in domains where physical laws are not yet defined.

\cite{nayek2021spike} focused on using spike-and-slab priors for discovering governing PDEs of
motion in nonlinear systems. The study used BLR for variable selection and parameter estimation
related to these governing equations.

In this paper, we investigate methods for extracting PDEs from raw datasets. We explore several
techniques, including Sparse Identification of Nonlinear Dynamics (SINDy), Least Absolute Shrinkage and Selection Operator (LASSO), Bayesian LASSO (B-LASSO), surrogate NNs, and SR. Our focus is on extracting
suitable PDEs for an MTS analysis and applying PINNs and B-PINNs for a regression task. This
approach is particularly applicable when dealing with raw data that lacks additional information
about its source or underlying dynamics. We then compare the results with and without the extracted
PDEs to evaluate any improvements in prediction accuracy for the MTS.

The contributions of this paper are outlined as follows: In Section \ref{S0}, we
explain the basic assumptions underlying PDEs. Section \ref{S1} presents various methods for
extracting PDEs from raw datasets. In Section \ref{S2}, we provide a brief overview of PINNs
through an algorithm. Section \ref{S3} discusses Bayesian versions of PINNs in conjunction with a
BLR model. A real MTS dataset is analyzed in Section \ref{S4} using the methods presented in this
paper. Finally, insights from the article are discussed in Section \ref{S5}.

\section{Differential Equations}\label{S0}
PDEs are mathematical equations that involve unknown functions and their
partial derivatives. They provide quantitative descriptions of a wide range
of phenomena across various fields, including physics, engineering, biology,
finance, and social sciences. PDEs typically involve multiple independent
variables and are essential for modeling dynamic systems where changes occur
over both time and space, \cite{tadmor2012review}.

On the other hand, ODEs involve functions of a single independent variable
and their derivatives, while PDEs incorporate functions of multiple
independent variables and their partial derivatives. The key distinction is
that ODEs deal with one-dimensional dynamics, whereas PDEs address
multi-dimensional phenomena.

An example of a first-order Ordinary Differential Equation (ODE) is $ \frac{dy}{dt} = ky $, where $ y $ is a
function of the single independent variable $ t $, and $ k $ is a constant.
This equation describes exponential growth or decay.

A well-known second-order PDE is the heat equation: $ \frac{\partial
\mathsf{T}}{\partial \mathsf{t}} = k \frac{\partial^2 \mathsf{T}}{\partial x^2} $, where $ \mathsf{T}(x,\mathsf{t}) $
represents the temperature at spatial location $ x $ and time $ \mathsf{t} $, with $ x
$ as the spatial independent variable and $ \mathsf{t} $ as the time variable,
\cite{zwillinger2021handbook}.

Dynamic systems refer to phenomena that evolve over time according to
specific rules or equations, as illustrated by the two examples above. These
systems can be linear or nonlinear, deterministic or stochastic, and they are
crucial for understanding complex behaviors in various applications, such as
climate modeling, population dynamics, and fluid mechanics.

\cite{peitz2024distributed} introduced a convolutional framework that
streamlines distributed reinforcement learning control for dynamical systems
governed by PDEs. This innovative approach simplifies high-dimensional
control problems by transforming them into a multi-agent setup with
identical, uncoupled agents, thereby reducing the model's complexity. They
demonstrated the effectiveness of their methods through various PDE examples,
successfully achieving stabilization with low-dimensional deep deterministic
policy gradient agents while using minimal computing resources.

\section{Extraction Methods of PDEs}\label{S1}
In this section, we explain several methods that are useful for constructing
PDEs from raw datasets. Some of these methods, such as SINDy and symbolic
regression, are well-known; however, they are not always successful across
all case studies. Additionally, surrogate NNs can be employed to generate
PDEs, but it is crucial that the NNs perform well on the unobserved test set
to ensure that its derivatives are reliable for building PDEs.

In this study, we have chosen Convolutional Neural Network (CNN) and Temporal Convolutional Network (TCN) architectures for the surrogate NNs
due to their accuracy in predictions. The constructed PDEs are local
approximations and, therefore, are not considered true PDEs. A larger dataset
can enhance the accuracy of these approximations, extending their validity to
a broader range of conditions and potentially revealing finer-scale features.

However, there are no guarantees that the approximation converges to the true
underlying PDE, if such a PDE exists. Even with a large training set, the
approximation may still overlook important aspects of the system's behavior
or overfit to noise in the data. Thus, while a larger dataset is
advantageous, it cannot transform a local approximation into a globally
accurate or true PDE; it can only improve the local approximation.

\subsection{SINDy}\label{S11}
SINDy is a powerful framework designed to uncover governing equations from data, especially in the
context of dynamical systems. Its primary goal is to identify the underlying equations that dictate
a system's behavior by making use of observed data. This method is particularly beneficial when the
governing equations are not known beforehand, which makes it a valuable asset in fields like fluid
dynamics, biology, and control systems. SINDy, introduced by \cite{brunton2016discovering}, employs
sparse regression alongside a library of candidate functions. This approach allows it to pinpoint
the essential terms in complex system dynamics, effectively capturing the key features needed to
accurately represent the data. Here is the breakdown of the SINDy algorithm to extract PDEs.

\cite{fasel2022ensemble} introduced an ensemble version of the SINDy algorithm, which improves the
robustness of model discovery when working with noisy and limited data. They demonstrated its
effectiveness in uncovering PDEs from datasets with significant measurement noise, achieving
notable improvements in both accuracy and robustness compared to traditional SINDy methods.
\cite{schmid2024ensemble} utilized the weak form of SINDy as an extension of the original algorithm
to discover governing equations from experimental data obtained through laser vibrometry. They
applied the SINDy to learn macroscale governing equations for beam-like specimens subjected to
shear wave excitation, successfully identifying PDEs that describe the effective dynamics of the
tested materials and providing valuable insights into their mechanical properties.

\textbf{SINDy Algorithm:}

\begin{enumerate}
  \item Prepare the system data, including measurements such as spatial and temporal elements of
      the state variable $X \in \mathbb{R}^{n \times d}$ for inputs and $Y \in \mathbb{R}^{n
      \times p}$ for outputs. Here, $n$ represents the number of samples, $d$ denotes the number
      of input features, and $p$ indicates the number of target variables. In the presence of
MTS, account for time steps, when preparing the output variables. For instance, the output
variable $Y$ should be shifted by $T$ samples from the input $X$ corresponding to the same
variables. This means that the first $T$ samples of the common variables in $X$ should be removed
to build the outputs. To simplify notation, we assume that the number of samples in $X$ and $Y$
is equal to $n$.

      In some applications, there might be only one dynamic variable (e.g., elapsed time $t$). In
      such cases, the input $X$ includes both dynamic and static variables, while the output $Y$
      consists of the same data as $X$ excluding the dynamic variable, shifted by at least one
      lag. Thus, $d = p + 1$ when there is only one dynamic variable present. For more
      clarification, one example is like:
\begin{equation*}
\begin{split}
X &= (\mathcal{T} \, (\text{time}), X_1 \, (\text{temperature}), X_2 \, (\text{moisture})), \\
Y &= (Y_1 \, (\text{temperature}), Y_2 \, (\text{moisture})),
\end{split}
\end{equation*}
      where $\mathcal{T}$ is the dynamic variable.
  \item Calculate the necessary derivatives $\dot{Y}\in \mathbb{R}^{n\times (d'\times p)}$ of the
      output variables with respect to the dynamic variables:
$$
\dot{Y}=\left( \frac{\partial Y}{\partial X_1},\cdots, \frac{\partial Y}{\partial X_{d'}}\right),
 \,\text{where } \frac{\partial Y}{\partial X_i} = \left(\frac{\partial Y_1}{\partial X_1},\cdots,
  \frac{\partial Y_p}{\partial X_{d'}}  \right), \text{ for } i=1,\cdots,d',
$$
where $\frac{\partial Y_j}{\partial X_i}$ for $j=1,\cdots,p$ is defined:
\begin{equation*}\label{NumericalDerivative}
  \frac{\partial Y_j}{\partial X_i} = \frac{Y_j(X_i+\Delta X_i)-Y_j(X_i)}{\Delta X_i}.
\end{equation*}
In this definition, $d'$ can be replaced by $d$ to take the derivatives with respect to all
  inputs if desired. This may involve numerical differentiation techniques to estimate the
  derivatives from the collected data. In this context, $d' \leq d$ represents the total number
  of dynamic variables.
  \item Create a library of potential basis functions for the state variables $\Theta(X)$, including polynomial terms, interactions between variables, trigonometric functions, and other relevant functions that may describe the dynamics of the system. For more complex models, one might also incorporate the derivatives of $X$ into the potential function, resulting in $\Theta(X, \dot{X})$.
  \item Formulate the sparse regression problem $\dot{Y} = \mathbf{c}\odot\Theta(X)$, where
      $\mathbf{c}$ is the coefficient vector to estimate, and $\odot$ is the Hadamard product.
  \item Optionally, use LASSO to regularize the coefficient vector. This step helps prevent the
      coefficients from becoming excessively large and promotes sparsity by shrinking some
      coefficients to zero. LASSO is a popular method for sparse regression as it adds a penalty
      term to the loss function that encourages sparsity. We refine the coefficient $\mathbf{c}$
  by minimizing:
  $$
  \min_{\mathbf{c}} \left( \|\dot{Y} - \mathbf{c}\odot \Theta(X)\|^2 + \lambda \|\mathbf{c}\|_1 \right),
  $$
  where $\|\cdot\|_1$ and $\|\cdot\|^2$ are the $L_1$ and $L_2$ norms, respectively.
  \item Select the active terms from the regression results that have non-zero coefficients.
  \item Assemble the identified terms into a mathematical model, typically in the form of a PDE.
  \item Validate the constructed model against the original data and select the equations that
      accurately describe the data.
\end{enumerate}

\subsection{LASSO}\label{S12}
As discussed in Subsection \ref{S11}, LASSO, introduced by \cite{tibshirani1996regression}, is a
robust regression technique particularly effective for extracting PDEs from data, especially within
the framework of PINNs. This method is valued for its dual capability of variable selection and
regularization, which allows for the most relevant feature identification while mitigating the
overfitting risk. LASSO achieves this by incorporating a penalty equal to the absolute value of the
coefficients into the loss function. This approach promotes sparsity in the model, effectively
reducing the number of variables retained in the final equation. Such sparsity is vital when
dealing with complex systems governed by PDEs, as it aids in pinpointing the most significant terms
from a potentially large pool of candidate variables derived from the data.

\cite{zhan2024physics} investigated the physics-informed identification of PDEs using LASSO
regression, particularly in groundwater-related scenarios. Their work illustrated how LASSO can be
effectively combined with dimensional analysis and optimization techniques, leading to improved
interpretability and precision in the resulting equations. Additionally, \cite{ma2024incorporating}
examined a variant version, sequentially thresholded least squares LASSO regression, within PINNs
to tackle inverse PDE problems. Their findings highlighted the method's ability to enhance
parameter estimation accuracy in complex systems. Their experiments on standard inverse PDE
problems demonstrated that the PINN integrated with LASSO significantly outperformed other
approaches, achieving lower error rates even with smaller sample data. The LASSO algorithm is
summarized in the following for better understanding of the concept.

\textbf{LASSO Algorithm:}

\begin{enumerate}
  \item Perform same as the SINDy Algorithm to define $\mathbf{X} = (X, \dot{X})\in \mathbb{R}^{n\times ( (p+1) \times d)}$. In this algorithm, the derivatives are obtained not only with respect to the dynamic variables, but also for other features of the input set.
  \item Formulate the LASSO regression model as:
  $$
  \min_{\mathbf{c}} \left( \|\dot{Y} - \mathbf{c} \odot\mathbf{X}\|^2 + \lambda \|\mathbf{c}\|_1 \right).
  $$
  \item Choose a range of $\lambda$ values to test (e.g., $0.01$, $0.1$, $1$, $10$, $100$).
  \item Implement $k$-fold cross-validation by splitting the data into $k$ subsets. Use each
      subset as the validation set and the remaining $k-1$ subsets as the training set.
  \item Fit the LASSO model to the training set using the current $\lambda$.
  \item Calculate the Mean Squared Error (MSE) or Mean Absolute Error (MAE) on the validation set and average them across all $k$ folds for each
      $\lambda$.
  \item Using the optimal $\lambda$ found from cross-validation, fit the LASSO model to the
      entire dataset $X$ and $Y$ to find $\mathbf{c}$. For simplicity, one can skip steps $3$ and $4$ and use a predetermined value for $\lambda$.
  \item Formulate the extracted function as a PDE based on the non-zero coefficients.
  \item Validate the extracted PDE against a separate dataset to ensure its predictive
      capability.
\end{enumerate}

\subsection{B-LASSO}\label{S13}

The Bayesian framework is particularly beneficial for dealing with limited data by incorporating
prior beliefs about the parameters. B-LASSO, \cite{park2008bayesian}, is a statistical method that
combines the principles of Bayesian inference with the LASSO regression technique, allowing us to
leverage the advantages of both approaches. This method provides uncertainty quantification and is
robust against overfitting, particularly in high-dimensional settings. It is especially useful for
regression problems where the number of predictors is large relative to the number of observations
or when there is multicollinearity among the predictors. In the Bayesian approach, we treat the
model parameters, known as coefficients, as random variables and specify prior distributions for
them. In addition to estimating the coefficients, B-LASSO also estimates the hyperparameter
$\lambda$ to control the strength of the LASSO penalty, assigning a prior distribution to $\lambda$
as well.

Many papers used B-LASSO for regression modeling. For example, \cite{chen2021partially} utilized
the B-LASSO for regression models and covariance matrices to introduce a partially confirmatory
factor analysis approach. They tackled the challenges of merging exploratory and confirmatory
factor analysis by employing a hierarchical Bayesian framework and Markov Chain Monte Carlo (MCMC) estimation, showcasing the
effectiveness of their method through both simulated and real datasets. In the following, we
explain the B-LASSO algorithm that is employed in this paper.

\textbf{B-LASSO Algorithm:}

\begin{enumerate}
  \item Follow step $1$ to $3$ in the SINDy Algorithm to prepare the data and define the design matrix.

  \item Formulate the B-LASSO model as follows:
  $$
  Y \sim \mathcal{N}\left(\bm{\beta} \odot\Theta(\bm{X}), \sigma^2 I_{(p+1)d} \right), 
  $$
  $$
  \bm{\beta} \sim \mathcal{L} \left( \bm{\mu}_{\beta}, \bm{\lambda} \right), \quad \bm{\lambda}
  \sim \mathcal{G}\left(\bm{\alpha}_{\lambda}, \bm{\beta}_{\lambda} \right), \quad \sigma \sim
  \mathcal{HC}\left( \mu_{\sigma}, \sigma_{\sigma} \right),\quad \sigma\geq\mu_{\sigma},
  $$
  where $\mathbf{X}=(X,\dot{X})$,
  $\bm{\beta} = (\beta_1, \ldots, \beta_{(p+1)d})$ represents the regression coefficients,
  $\bm{\lambda} = (\lambda_1, \ldots, \lambda_{(p+1)d})$ denotes the positive real regularization
  parameters, and $\sigma$ is the model noise factor. The hyperparameters $\bm{\mu}_{\beta} =
  (\mu_{\beta_1}, \ldots, \mu_{\beta_{(p+1)d}})$, $\bm{\alpha}_{\lambda} = (\alpha_{\lambda_1},
  \ldots, \alpha_{\lambda_{(p+1)d}})$, $\bm{\beta}_{\lambda} = (\beta_{\lambda_1}, \ldots,
  \beta_{\lambda_{(p+1)d}})$, $\mu_{\sigma}$, and $\sigma_{\sigma}$ can be set based on available
  data. The symbols $\mathcal{N}$, $\mathcal{L}$, $\mathcal{G}$, and $\mathcal{HC}$ represent the
  Normal, Laplace, Gamma, and Half-Cauchy distributions, respectively. For simplicity,
  $\Theta(\cdot)$ can be identity function.

  \item Estimate the posterior distribution using a sampling method such as MCMC. The posterior distribution of the parameters $\bm{\beta}$, $\bm{\lambda}$, and $\sigma$ given the data $Y$ and $\Theta(\bm{X})$ is expressed as:
  $$
  P(\bm{\beta}, \bm{\lambda}, \sigma \mid Y, \Theta(\bm{X})) \propto P(Y \mid \Theta(\bm{X}), \bm{\beta}, \sigma) P(\bm{\beta} \mid \bm{\lambda}) P(\bm{\lambda}) P(\sigma).
  $$

  \item Approximate the coefficients $\bm{\beta}$ using the posterior distribution samples:
  $$
  \hat{\bm{\beta}} = E(\bm{\beta} \mid Y, \Theta(\bm{X})) \approx \frac{1}{N M}\sum_{i=1}^{N} \sum_{j=1}^{M} \hat{\bm{\beta}}_{ij},
  $$
  where $\hat{\bm{\beta}}_{ij}$ represents the $i$-th sample of $\bm{\beta}$ from the $j$-th chain, $N$ is the number of draws per chain, and $M$ is the number of chains. In this context, the highest density interval of the coefficients can be used to measure uncertainty for a specified confidence level, such as $95\%$. The HDI is a Bayesian credible interval that contains the most probable values of the posterior distribution.

  \item Construct PDEs based on the non-zero coefficients in the form $Y \approx \hat{\bm{\beta}}
      \odot\Theta(\bm{X})$.

  \item Evaluate the extracted PDEs using unseen data and select the equations that demonstrate high accuracy.
\end{enumerate}

\subsection{CNN}\label{S14}

CNNs are well-known for their ability to handle spatial relationships in various types of data, including images, videos, and even speech. They are also effective for MTS data. Typically, one-dimensional convolution is used for MTS, but two-dimensional convolution can be applied when the data has two-dimensional features.
One of the key advantages of CNNs is their ability to automatically learn patterns and features from raw data, which reduces the need for manual feature engineering. They achieve this by using filters that share parameters across different parts of the input, resulting in lower memory usage and improved performance.
CNNs are capable of learning hierarchical representations, where lower layers capture simple features and deeper layers identify more complex patterns and abstractions. However, they usually require a large amount of training data to perform well, which can be a barrier in some applications. Additionally, CNNs are sensitive to small changes in the input data, such as noise, translation, or rotation, which can impact prediction accuracy.
This sensitivity can lead to overfitting, especially when working with small datasets. To mitigate this issue, techniques like dropout, data augmentation, and regularization can be beneficial.

Recent research has explored various versions and combinations of CNNs to enhance machine learning
efficiency across different problems. For instance, \cite{liu2025cnn} introduced an innovative
approach for airfoil shape optimization that integrates CNNs to extract and compress airfoil
features, PINNs for assessing aerodynamic performance, and deep reinforcement learning for identifying optimal solutions.
This integrated methodology successfully reduces the design space and improves the lift-drag ratio
while tackling the challenges of high-dimensional optimization and performance evaluation.

One important aspect to consider when using surrogate NNs, such as CNNs and TCNs, for extracting
PDEs is that the network must perform well on unseen datasets. This means that the surrogate NN
should effectively learn a function that closely approximates the true relationship between the
inputs and outputs. Consequently, by differentiating the network’s output with respect to its
inputs, we can derive the approximate true PDEs. To construct PDEs, it is beneficial to incorporate
dynamic variables as inputs to the network. This allows the network to be differentiated with
respect to these dynamic inputs after training, facilitating the construction of PDEs. It is
important to note that while the final equations may not always be true PDEs—since they are defined
based on the dataset's domain—they serve as local approximations of the true PDEs. If the training
dataset is comprehensive, the final equations closely approximate the true PDEs. This is the
convolutional surrogate NN algorithm designed to extract PDEs from data.

\textbf{CNN Algorithm:}

\begin{enumerate}
  \item Provide the input features $X \in \mathbb{R}^{n \times T \times d}$ and the target variables $Y$, where $T$ is the number of time steps, known as lags. Since we are focusing on MTS, $T$ is crucial for ensuring that the inputs and outputs are derived from the same dataset. For example, $T = 10$ means that $10$ previous time steps are used to predict the value at time $T + 1$.

  \item Apply one-dimensional convolutional layers with filters $W_l \in \mathbb{R}^{m \times d}$ to the $l$-th layer, where $m$ is the kernel size. The output of the $l$-th convolutional layer is:
  $$
  H_l = \text{ReLU}(W_l * H_{l-1} + b_l),
  $$
  where $*$ denotes the convolution operation, $H_{l-1}$ is the output of the previous layer, $b_l$ is the bias term, and $\text{ReLU}$ is the activation function.

  \item Use max-pooling to reduce the dimensionality of the feature maps:
  $$
  P_l = \text{MaxPool}(H_l).
  $$

  \item Flatten the output of the final convolutional layer and pass it through a fully connected layer to produce the predicted outputs $\hat{Y} \in \mathbb{R}^{n \times p}$:
  $$
  \hat{Y} = W_f \cdot \text{Flatten}(P_k) + b_f,
  $$
  where $W_f$ and $b_f$ are the weights and biases of the fully connected layer, and $\cdot$
  denotes matrix multiplication.

  \item Train the model using the MSE loss function. To obtain accurate derivatives, it is essential to have a well-trained network, as this ensures the reliability of the derivatives produced afterward.

  \item Compute the derivatives of the predicted outputs $\hat{Y}$ with respect to the input features $X$ using automatic differentiation:
  $$
  \frac{\partial \hat{Y}}{\partial X} = \nabla_X \hat{Y}.
  $$

  \item Organize the derivatives into a matrix $D \in \mathbb{R}^{n \times T \times (p \times d)}$, where each entry $D_{i,t,k}$ represents the derivative of the $j$-th output variable $\hat{Y}_{i,j}$ with respect to the $k$-th input feature $X_{i,t,k}$ at time step $t$ for the $i$-th sample:
  $$
  D_{i,t,k} = \frac{\partial \hat{Y}_{i,j}}{\partial X_{i,t,k}}.
  $$
  \item Perform a correlation analysis between the derivatives and the target variables to identify the most significant terms in the PDEs. Compute the correlation matrix to quantify the relationships.

  \item Select the derivatives with the highest correlation (e.g., correlation greater than $0.5$) to the target variables. These derivatives are considered candidate terms for the PDEs. For example, to construct a PDE for a specific output feature with respect to a specific input feature, such as $\frac{\partial \hat{Y}_1}{\partial X_1}$, we select the most relevant derivatives and state variables associated with the desired derivative.

  \item Approximate the relationship between the significant derivatives and the target variables using a polynomial regression model. The resulting polynomial equation represents the extracted PDE:
  $$
  \frac{\partial \hat{Y}}{\partial X} = \beta_0 + \sum_{i=1}^{q} \beta_i \Phi_i,
  $$
  where $\Phi$ represents polynomial features of degree $q$ constructed from the significant derivatives and state variables.

  \item Evaluate the accuracy of the extracted PDEs and select those with the highest predictive performance.
\end{enumerate}

\subsection{TCN}\label{S15}

TCNs are a type of NN architecture specifically designed for trajectory data and powerful in handling temporal information. They utilize causal convolutions, meaning that the output at any given time step depends only on current and past input values. This characteristic makes TCNs particularly well-suited for tasks such as MTS forecasting, speech synthesis, and other forms of temporal sequence analysis.
By employing dilated convolutions, TCNs can effectively expand their receptive fields, allowing them to capture long-range dependencies without relying on recurrent connections. This design also enables TCNs to process data across time steps in parallel, resulting in faster training and more stable gradients compared to recurrent NNs.
However, while TCNs provide faster training, using very large kernel sizes or high dilation rates for long sequences can increase computational costs and memory usage due to the larger number of parameters involved. The choice of kernel size, stride, and dilation is crucial, as these factors can significantly impact performance and should be carefully selected based on the characteristics of the input data.
Additionally, TCNs can be prone to overfitting, especially when trained on small datasets, which necessitates the use of regularization techniques to improve generalization.

In a recent article, \cite{perumal2023temporal}, the authors explored the use of TCNs for rapidly inferring thermal histories in metal additive manufacturing. They demonstrated that TCNs can effectively capture nonlinear relationships in the data while requiring less training time compared to other deep learning methods. The study highlighted the potential advantages of integrating TCNs with PINNs to enhance modeling efficiency in complex manufacturing contexts.
Next, we outline the algorithm used for the TCN model presented in this work.

\textbf{TCN Algorithm:}

\begin{enumerate}
  \item Follow step $1$ of the CNN Algorithm.

  \item Apply a TCN to the input data. The TCN consists of multiple dilated convolutional layers with causal padding to preserve the temporal order of the data. The output of the $l$-th dilated convolutional layer is given by:
  $$
  H_{l_d} = \text{ReLU}(W_l *_{r_l} H_{l-1} + b_l),
  $$
  where $*_{r_l}$ denotes the causal convolution operation with a dilation rate $r_l$ that increases exponentially with each layer. To ensure that the model does not utilize future information for predictions, causal padding is applied. For a kernel of size $m$ and a dilation rate $r_l$, the input is padded with $(m-1)r_l$ zeros on the left side.

  \item Add residual connections to improve gradient flow and stabilize training. For the $l$-th layer, the residual connection is computed as:
  $$
  h(H_{l-1}) = W_{\text{res}} * H_{l-1} + b_{\text{res}},
  $$
  where $W_{\text{res}}$ and $b_{\text{res}}$ are the residual weights and bias. The final output of the layer is:
  $$
  H_l = H_{l_d} + h(H_{l-1}).
  $$

  \item Perform steps $4$ to $11$ of the CNN Algorithm.
\end{enumerate}

\subsection{Symbolic Regression (SR)}\label{S16}

SR is a sophisticated method used in Machine Learning (ML) and artificial intelligence to identify mathematical
expressions that best fit a given dataset. Unlike traditional regression techniques that rely on
predefined models, SR seeks to discover both the model structure and parameters directly from the
data. It employs techniques such as genetic programming, NNs, and other optimization algorithms to
derive these mathematical expressions such as complex and nonlinear relationships. The process
involves generating a variety of candidate functions and using optimization methods to evaluate
their performance against the data. This approach can reveal underlying physical laws or
relationships that standard regression techniques might overlook, providing valuable insights into
the system being studied. However, the search space for potential mathematical expressions can be
vast, resulting in high computational costs and longer training times, especially with complex
datasets. There is also a risk of overfitting, particularly if the search space is not
well-regularized or if the dataset is small. Achieving optimal results often requires careful
tuning of several hyperparameters.

SR has applications across various fields, including data science for uncovering relationships
during exploratory data analysis, engineering for deriving mathematical models of complex systems,
biology for understanding biological processes or gene interactions, and finance for modeling
market behaviors and predicting trends. \cite{changdar2024integrating} introduced a hybrid
framework that merged ML with SR to analyze nonlinear wave propagation in arterial blood flow. They
utilized a mathematical model along with PINNs to solve a fifth-order nonlinear equation,
optimizing the solutions through Bayesian hyperparameter tuning. This approach led to highly
accurate predictions, which were further refined using random forest algorithms. Additionally, they
applied SR to extract interpretable mathematical expressions from the solutions generated by the
PINNs.

\textbf{SR Algorithm:}

\begin{enumerate}
  \item Prepare the relevant input features $X \in \mathbb{R}^{n \times d}$ and target features $Y \in \mathbb{R}^{n \times p}$.
  \item Compute the derivative matrix $D \in \mathbb{R}^{n \times (d \times p)}$ of the output variables with respect to the input variables.
  \item Utilize a SR package, such as gplearn (a Python library for SR based on genetic
      programming), to approximate the relationship between the derivatives and the target
      variables.
  \item Fit a symbolic regressor $\mathcal{S}$ to the training data $D$ to approximate the relationship:
  $$
  \mathcal{S}: D \rightarrow Y.
  $$
  The symbolic regressor searches for a mathematical expression $\Theta$ that minimizes the error:
  $$
  \Theta^* = \arg \min_\Theta \|Y - \Theta(D)\|^2.
  $$
  \item Extract the best-performing equation $\Theta^*$ from the symbolic regressor $\mathcal{S}$. For example, one PDE may have the form:
  $$
  \frac{\partial Y_1}{\partial X_1} = \Theta^* \left( \frac{\partial Y}{\partial X_{-1}} \right),
  $$
  where $Y_1$ is the first target feature, $X_1$ is the first input feature, and $X_{-1}$ represents all input features except the first one.
  \item Evaluate the extracted PDEs and select those with high accuracy.
\end{enumerate}

\section{PINN}\label{S2}
PINNs, introduced by \cite{raissi2019physics}, represent a powerful technique that integrates data-driven ML with fundamental physical principles, remaining it with low computational cost. Unlike traditional NNs, which rely solely on data for training, PINNs incorporate known physical laws expressed as PDEs and boundary conditions into the loss functions. This integration acts as a regularization mechanism, enabling PINNs to achieve higher accuracy, especially in scenarios with limited or noisy data. By leveraging the underlying physics, PINNs can generalize better and make predictions in regions where data is sparse or unavailable.

Since $2019$, numerous research papers explored PINNs with various NN architectures and
applications. For instance, \cite{hu2024physics} discussed how PINNs effectively integrated
imperfect and sparse data. They also addressed how PINNs tackled inverse problems and improved
model generalizability while maintaining physical acceptability. This was particularly relevant for
solving problems related to PDEs that governed the behavior of solid materials and structures.
Additionally, their paper covered foundational concepts, applications in constitutive modeling, and
the current capabilities and limitations of PINNs. Here is the structure of the applied PINNs in
this article. One can exclude steps $3$ and $5$ of the PINN Algorithm to adapt it into a typical
NN.

\textbf{PINN Algorithm:}

\begin{enumerate}
  \item Prepare a new dataset of inputs $X$ and outputs $Y$, similar to step $1$ in the SINDy Algorithm.
  \item Create a NN architecture aligned with the number of inputs and outputs according to the data.
  \item Define the physical loss function $\mathcal{L}_P$ based on the extracted PDEs. For example, if a PDE is extracted as
  $$\frac{\partial Y_j}{\partial X_i} = f\left( \frac{\partial Y}{\partial X_{-i}} \right),$$
  where $f(\cdot)$ is a non-linear function, $i \in \{1, \ldots, d'\}$, and $j \in \{1, \ldots, p\}$, then the loss function is defined by the MSE:
  $$
  \mathcal{L}_P = \sum_{s=1}^{n} \sum_{i=1}^{d'} \sum_{j=1}^{p} a_j \left\| \frac{\partial Y_{js}}{\partial X_{is}} - f\left( \frac{\partial Y_s}{\partial X_{-i\, s}} \right) \right\|^2,
  $$
  where $a = (a_1, \ldots, a_{p})$ is a vector of zeros and ones. The values of $a_j$ indicate whether the corresponding derivative equations are included in the loss function based on the PDE extraction. If $a_j=1$, then $\frac{\partial Y_j}{\partial X_i}$ contributes to the loss function; conversely, $a_j=0$ indicates that no equation is defined for $\frac{\partial Y_j}{\partial X_i}$, and it is excluded from the loss.

  \item Determine the data loss function $\mathcal{L}_D$:
  $$
  \mathcal{L}_D = \sum_{s=1}^{n} \sum_{j=1}^{p} \left\| Y_{js} - \hat{Y}_{js} \right\|^2.
  $$

  \item Combine the physics loss and data loss into a total loss function $\mathcal{L}_T$:
  \begin{equation}\label{PINN_Loss}
    \mathcal{L}_T = \mathcal{L}_P + \mathcal{L}_D.
  \end{equation}
  Although MSE is used in this algorithm, other loss functions can also be applied.

  \item Use the Adam optimizer or any other suitable optimizer to minimize $\mathcal{L}_T$ with respect to the network weights and biases.

  \item Train the model for a fixed number of epochs.

  \item Evaluate the trained model on the validation and test sets using the total loss function. If necessary, refine or tune the hyperparameters to achieve better results.
\end{enumerate}

\section{Bayesian Computations}\label{S3}
Bayesian computational models are well-known for their ability to measure the uncertainty of
variables and perform effectively on noisy data. However, their high computational cost can be a
significant drawback, making them less accessible for researchers with limited resources. In this
section, we introduce a Bayesian model, called BLR, and a B-PINN that we utilize in our work.

\subsection{Physics-Informed Bayesian Linear Regression (PI-BLR)}\label{S31}
BLR provides a robust framework for statistical modeling by integrating prior beliefs and
accounting for uncertainty in predictions. Given our interest in incorporating PDEs as a negative
potential component in the model, the simplicity of a linear model makes it easy to derive the
necessary derivatives. This approach not only improves predictive performance but also supports
more informed decision-making across various fields, including finance and healthcare.

\cite{fraza2021warped} introduced a warped BLR framework, specifically applied to data from the UK
Neuroimaging Biobank. The paper emphasized the benefits of this approach, such as enhanced model
fitting and predictive performance for various variables, along with the capacity to incorporate
non-Normal data through likelihood warping. This method had less computationally intensive than
Normal process regression, as it eliminated the need for cross-validation, making it well-suited
for large and sparse datasets. \cite{gholipourshahraki2024evaluation} applied BLR to prioritize
biological pathways within the genome-wide association study framework. The advantages of employing
BLR included its capacity to reveal shared genetic components across different phenotypes and
enhanced the detection of coordinated effects among multiple genes, all while effectively managing
diverse genomic features.

In this subsection, we integrate PDEs into the BLR model to introduce physical insights into the
framework, called PI-BLR. The algorithm is outlined as follows. For a standard BLR model, simply
omit steps $3$, $5$, and $6$.

\textbf{PI-BLR Algorithm:}

\begin{enumerate}
  \item Prepare the input variable $X \in \mathbb{R}^{n \times d}$ and the output target $Y \in
      \mathbb{R}^{n \times p}$, similar to the SINDy Algorithm.

  \item Define the priors and the Normal likelihood for the observed outputs:
  $$
  Y \sim \mathcal{N}\left(\bm{\beta} \odot X, \sigma^2 I_{p} \right),
  $$
  $$
  \bm{\beta} = \bm{\lambda} \odot \epsilon,
  \quad \bm{\lambda} \sim \mathcal{G}\left(\bm{\alpha}_{\lambda}, \bm{\beta}_{\lambda}\right), \quad \epsilon \sim \mathcal{N}\left(0,\sigma^2_{\epsilon}\right),
  \quad \sigma \sim \mathcal{HC}\left(\mu_{\sigma}, \sigma_{\sigma}\right),
  $$
  where $\sigma^2_{\epsilon}$, $\mu_{\sigma}$, and $\sigma_{\sigma}$ are hyperparameters.
  The term
  $\epsilon$ is included to enhance sampling efficiency, while $\bm{\beta}$ represents
the regression coefficients with a non-centered parameterization for stable sampling. The scale
parameters $\bm{\lambda}$ enforce sparsity through Gamma priors.

  \item Specify the desired PDEs as in step 3 of the PINN Algorithm.

  \item Construct the posterior distribution of $\bm{\beta}$, $\bm{\lambda}$, and $\sigma$ given
      the data $X$ and $Y$:
  $$
  P(\bm{\beta}, \bm{\lambda}, \sigma \mid Y, X) \propto P(Y \mid X, \bm{\beta}, \sigma) P(\bm{\beta} \mid \bm{\lambda}) P(\bm{\lambda}) P(\sigma).
  $$

  \item Set the PDEs to zero:
  $$
  \text{PDE}_{ji}(\bm{\beta}) = \frac{\partial Y_j}{\partial X_i} - f\left( \frac{\partial Y}{\partial X_{-i}} \right), \quad j=1,\cdots,p, \, i=1,\cdots,d.
  $$
  Note that $\text{PDE}_{ji}$ does not need to be defined for all $j$ and $i$. For undefined
  cases, we assume it to be zero. For simplicity, assume there are $p'$ PDEs, which we can denote
  as $\text{PDE} = (\text{PDE}_1, \cdots, \text{PDE}_{p'})$. The derivatives are inherently
  dependent on the coefficients $\bm{\beta}$, although this dependence may not be explicitly
  clear in the preceding formula. This relationship is derived based on the model established in
  the step 2. The PDE residuals act as soft constraints, weighted by $\sigma_{\text{PDE}}$, which
  can be interpreted as assuming the PDE residuals follow a zero-mean Normal distribution:
  $$
  \text{PDE} \sim \mathcal{N}\left(0, \sigma^2_{\text{PDE}}\right).
  $$
  The logarithmic term for the PDEs is given by:
  $$
  \log P(\text{PDE} \mid \bm{\beta}) = -\frac{1}{2\sigma^2_{\text{PDE}}} \sum_{i=1}^{p'} \text{PDE}^2_i + \mathcal{C}.
  $$
  Here, $\log P(\text{PDE}\mid\bm{\beta})$ serves as a quadratic penalty, equivalent to a Normal
  prior on the PDE residuals.

  \item Rewrite the posterior distribution to incorporate the physics-based potential term,
      penalizing deviations from domain-specific PDEs:
  $$
  P(\bm{\beta}, \bm{\lambda}, \sigma \mid Y, X) \propto P(Y \mid X, \bm{\beta}, \sigma) P(\bm{\beta} \mid \bm{\lambda}) P(\bm{\lambda}) P(\sigma) P(\text{PDE} \mid \bm{\beta}).
  $$

  \item Find the maximum a posteriori estimation to initialize the parameters:
  $$
  \bm{\beta}_0, \bm{\lambda}_0, \sigma_0 = \arg \max_{\bm{\beta}, \bm{\lambda}, \sigma} \log P(\bm{\beta}, \bm{\lambda}, \sigma \mid Y, X),
  $$
  where $\bm{\beta}_0$, $\bm{\lambda}_0$, and $\sigma_0$ are the initialization parameter
  estimations.

  \item Sample from the full posterior using the No-U-Turn Sampler (NUTS), and specify appropriate values for the
      number of chains, tuning steps, number of returned samples, target acceptance rate, and
      maximum tree depth. It is worth mentioning that tuning steps, also known as burn-in, refer
      to the number of initial samples to discard before collecting the desired number of samples
      from each chain.

  \item Evaluate the model on an unseen dataset and adjust the hyperparameters as necessary.
\end{enumerate}

\subsection{B-PINN}\label{S32}

Bayesian methods are particularly robust in the presence of noisy or sparse data because they
incorporate prior knowledge and propagate uncertainty throughout the model. By treating NN weights
and biases as random variables, these methods estimate their posterior distribution based on the
available data and physical constraints, enabling effective uncertainty measurement. Integrating
Bayesian inference into PINNs naturally regularizes the model, helping to prevent overfitting and
improve generalization. By providing posterior distributions, B-PINNs offer valuable insights into
the uncertainty and reliability of predictions. However, it is important to recognize that Bayesian
inference techniques, such as MCMC and Metropolis sampling, can be computationally intensive,
particularly when addressing high-dimensional problems.

\cite{yang2021b} developed a B-PINN to address both forward and inverse nonlinear problems governed
by PDEs and noisy data. Utilizing Hamiltonian Monte Carlo and variational inference, the model
estimated the posterior distribution, enabling effective uncertainty quantification for
predictions. Their approach not only addressed aleatoric uncertainty from noisy data but also
achieved greater accuracy than standard PINNs in high-noise scenarios by mitigating overfitting.
The posterior distribution was used to estimate the parameters of the surrogate model and the PDE.
The PDE was incorporated into the prior to impose physical constraints during the training process.

There are numerous approaches to transform a PINN into a B-PINN. The paper by
\cite{mohammad2025bayesian} introduces a Bayesian physics-informed framework that enhances NN
training by integrating domain expertise and uncertainty quantification. Unlike traditional PINNs,
which impose physical laws as soft constraints, this approach formalizes physical equations and
prior knowledge within a probabilistic framework. The method constructs a loss function that
balances three key components: fitting observed data, adhering to physical laws, and regularizing
predictions based on prior estimates from a Bayesian perspective. In this paper, we utilize a BNN
and incorporate the extracted PDEs into the data loss function to form a kind of B-PINN. In the
following, we outline the B-PINN used here. For the corresponding BNN, we exclude steps $5$, which
involve the incorporation of PDEs.

\textbf{B-PINN Algorithm:}

\begin{enumerate}
  \item Start by following step $1$ of the SINDy Algorithm to load the input and output datasets.
  \item Define the initial layer and subsequent layers using a non-centered parameterization of
      weights:
\begin{align*}
  H_1 & = \tanh(X (cW_1) + b_1), \\
  H_l & = \tanh(H_{l-1} (cW_l) + b_l), \quad l=2, \cdots, L-1, \\
  \hat{Y} & = H_{L-1} (cW_L) + b_L,
\end{align*}
  where $c$ is a small scalar, such as $0.01$, to ensure the weights are non-centered, and $L$
  represents the total number of layers. All weights $\bm{W} = (W_1, \cdots, W_L)$ and biases
  $\bm{b} = (b_1, \cdots, b_L)$ are assumed to follow distributions $\mathcal{N}(0, \sigma^2_W)$
  and $\mathcal{N}(0, \sigma^2_b)$, respectively, with typical values being $\sigma^2_W = 1$ and
  $\sigma^2_b = 0.1$. The hyperbolic tangent function is omitted in the last layer for regression
  tasks.
  \item Assign a Half-Normal prior to the observational noise, represented as $\sigma \sim
      \mathcal{HN}(\sigma_{\sigma})$, where $\mathcal{HN}$ denotes the Half-Normal distribution.
      This prior, being lighter-tailed compared to Half-Cauchy, enhances identifiability in
      high-dimensional B-PINNs by penalizing large noise values.
  \item Construct the likelihood function based on the prediction $\hat{Y} \in \mathbb{R}^{n
      \times p}$:
      $$
      Y \sim \mathcal{N}(\hat{Y}, \sigma^2 I_p).
      $$
  \item Formulate the PDEs as outlined in steps $3$ and $5$ of the BLR Algorithm.
  \item Define the hierarchical Bayesian inference using either the PDEs or by omitting their
      current term to formulate a BNN:
\begin{align*}
  P(\bm{W}, \bm{b} \mid X, Y) \propto & \underbrace{P(X, Y \mid \bm{W}, \bm{b})}_{\text{Likelihood}} \underbrace{P(\bm{W}) P(\bm{b})}_{\text{Prior}} \underbrace{P(PDE \mid \bm{W}, \bm{b})}_{\text{Physics}} \\
\end{align*}
  \item Use Automatic Differentiation Variational Inference (ADVI) to approximate initial points for the parameters. ADVI is a type of variational
      inference aimed at approximating the true posterior distribution by a simpler,
      parameterized distribution, achieved by minimizing the Kullback-Leibler divergence between the two
      distributions.
  \item Sample from the approximated posterior distribution using NUTS with the hyperparameters
      defined in step $8$ of the BLR Algorithm.
  \item Evaluate the trained model to determine its reliability.
\end{enumerate}

\section{A Real World MTS Dataset}\label{S4}

We utilize the \textit{Household Electric Power Consumption} dataset from
\cite{electric_power_consumption}, which comprises $2,075,259$ samples recorded at one-minute
intervals over a period of nearly four years. The data spans from December $16$, $2006$, at $17:24$
to November $26$, $2010$, at $21:02$. Each sample includes seven measured variables alongside the
corresponding date and time. Although the dataset does not specify the exact location where the
measurements were taken, it provides detailed insights into the energy consumption patterns of a
single household.

We add a dynamic variable to the dataset by calculating the elapsed time in hours for each sample,
starting from $0$ for the first sample using the information in the date and time columns. To gain
a better understanding of the data, we calculate the correlations between each pair of variables.
We find that two variables have a correlation of exactly $1$, indicating that they are linear
functions of each other. In other words, knowing the value of one variable allows us to determine
the exact value of the other using a simple regression model to derive the linear relationship.

The dataset captures the following key electrical quantities, which we have named for easier
handling. The input variables are designated by $X$ indices, while the outputs are represented by
$Y$ indices. Since this is a MTS, some of the inputs and outputs may overlap, considering an
appropriate lag like $30$:

\begin{enumerate}
    \item \textbf{Elapsed Time} $X_1$: The time that has passed since the beginning of the data
        collection period, measured in hours. This variable enables us to analyze trends and
        patterns in energy consumption over time, facilitating time-based analyses and
        comparisons.

    \item \textbf{Global Active Power} $X_2$ and $Y_1$: The total active power consumed by the
        household, averaged over each minute (in kilowatts).

    \item \textbf{Global Reactive Power} $X_3$ and $Y_2$: The total reactive power consumed by
        the household, averaged over each minute (in kilowatts).

    \item \textbf{Voltage} $X_4$ and $Y_3$: The minute-averaged voltage (in volts).

    \item \textbf{Global Intensity}: The minute-averaged current intensity (in amperes). This
        variable is a linear function of Global Active Power, so we omit it from the dataset.

    \item \textbf{Sub-metering$_1$} $X_5$ and $Y_4$: Energy consumption (in watt-hours)
        corresponding to the kitchen, primarily attributed to appliances such as a dishwasher,
        oven, and microwave.

    \item \textbf{Sub-metering$_2$} $X_6$ and $Y_5$: Energy consumption (in watt-hours)
        corresponding to the laundry room, including a washing machine, tumble-drier,
        refrigerator, and lighting.

    \item \textbf{Sub-metering$_3$} $X_7$ and $Y_6$: Energy consumption (in watt-hours)
        corresponding to an electric water heater and air conditioner.
\end{enumerate}

Notably, the dataset contains approximately $1.25\%$ missing values, represented by the absence of
measurements between consecutive timestamps. We apply first-order spline interpolation to fill in the missing parts.
Additionally, the active energy consumed by appliances not covered by the
sub-metering systems can be derived using the formula:
$$
\text{Active Energy (Wh)} = X_2 \times \frac{1000}{60} - X_5 - X_6 - X_7.
$$

To enhance the effectiveness of ML methods, normalization is recommended to improve the learning
process. In this example, we first normalized the dataset and then applied the described method for
further analysis.

In the PDE extraction phase, the most effective method is the surrogate NN utilizing a TCN architecture. We employ two models and explain the differences between them.
Model $1$ is a standard TCN with three convolutional blocks, utilizing filters of sizes $64$ and $128$ and a kernel size of $3$. It features residual connections and dropout layers to reduce overfitting, and it is trained using the Adam optimizer.
Model $2$ enhances this foundation by incorporating dilated convolutions, which allow it to capture long-range dependencies more effectively. With dilation rates of $[1, 2, 4]$, it can analyze broader contexts without a significant increase in computational cost. Retaining the residual connections and dropout for stability, model $2$ is specifically trained to focus on temporal dependencies, making it more adept at recognizing patterns over extended periods.
The CNN model employs a one-dimensional CNN that consists of three convolutional blocks. Each block includes ReLU activation, max pooling, and dropout layers to effectively extract temporal patterns while minimizing the risk of overfitting. The training process utilizes the Adam optimizer along with MSE as the loss function. All networks are trained for $30$ epochs.

The method used in the surrogate NNs involves training the network on a large historical dataset and then utilizing the trained model to perform differentiation. The key aspect is that the networks must make accurate predictions. If they do, we can use the trained network to calculate derivatives. This implies that when the network accurately predicts unseen data, it identifies the unknown function of the input to generate the outputs. Consequently, we can differentiate this network, which represents a mathematical function of the inputs.

\begin{figure}[t] 
    \centering
        \includegraphics[width=\textwidth, height=8.5cm]{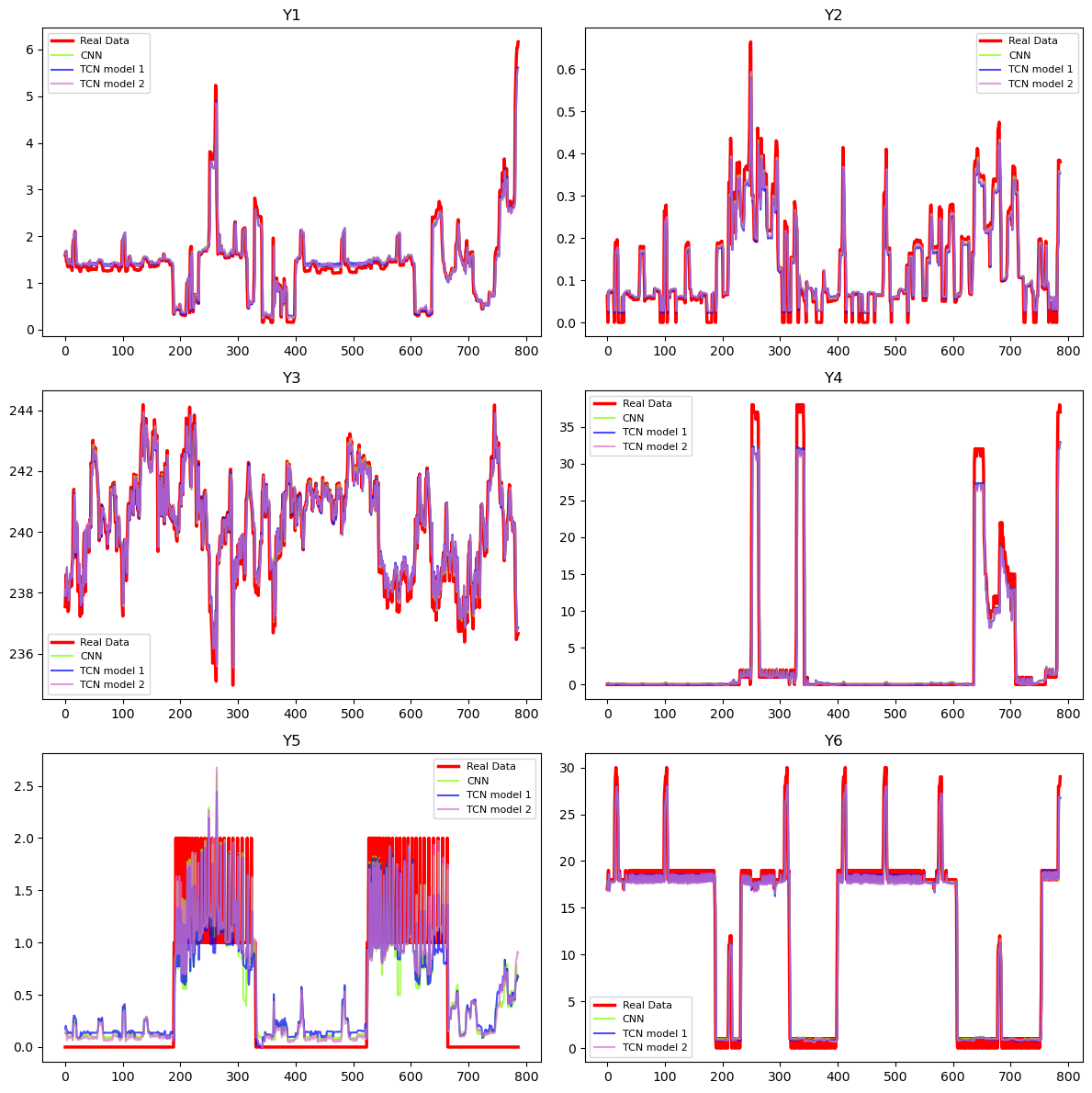} 
    \caption{Predictions of the state variables in PDE extraction phase.} 
    \label{Predictions_PDEextraction_Y}
\end{figure}

Table \ref{PDEextractionPhaseMetrics} presents the metrics of the trained network to assess its predictive performance. Figure \ref{Predictions_PDEextraction_Y} displays the corresponding predictions of the surrogate NNs. Based on the results, all three networks make good predictions of the variables, except for $Y_5$.
In the next step, we calculate the derivatives of the outputs with respect to all input variables in the train set. Then, we formulate PDEs with respect to the dynamic variable $X_1$ using other derivatives based on polynomials of degree $3$. The results of the surrogate NNs, along with those from other methods, are presented in Table \ref{PDEextractionPhaseMetricsderivatives}.
Since the metrics indicate significantly large derivatives for other methods, we plot the PDEs only for the surrogate NNs in Figure \ref{Predictions_PDEextraction_derivatives}, separately for each network. This is because each NN predicts the outputs using different mathematical functions, resulting in distinct derivatives for each network.

The $R^2$  values are predominantly negative or near zero, indicating poor model performance in Table \ref{PDEextractionPhaseMetricsderivatives}. This occurs because $R^2 = 1- SSR/SST$ becomes negative when the model’s prediction errors (SSR) exceed the natural variability in the data (SST). Such results suggest the model fails to capture meaningful patterns.
We only use the PDEs from the TCN models, either model $1$ or model $2$, despite the CNN showing some high values of $R^2$. The reason is that the prediction plots in Figure \ref{DerivaPredCNN} display significant variability, making the derivatives unreliable. In contrast, the predictions for $dY_2/dX_1$, $dY_3/dX_1$, and $dY_6/dX_1$ in model $1$, as well as for $dY_1/dX_1$ in model $2$, are very accurate. In model $1$, the $R^2$ for $dY_2/dX_1$ is $-0.4289$, largely due to one part of data that is significantly distant from the exact values.

\begin{table} 
    \centering
    \caption{Performance metrics in PDE extraction phase for the state variable predictions.}\label{PDEextractionPhaseMetrics}
    \begin{tabular}{|c|c|c|c|c|c|}
        \hline
        Phase & Methods & Variable & MSE  & MAE & $R^2$ \\ \hline
\multirow{18}{*}{\makecell{PDE Extractions \\ $1958991$ training \\ $19000$ validation \\ $788$ testing}} & \multirow{6}{*}{CNN} & $Y_1$ &
$0.0706$ & $	0.1383$ & $	0.8853$ \\
         & & $Y_2$ & $0.0021$ & $	0.0262$ & $	0.8483$ \\
         & & $Y_3$ & $0.4367$ & $	0.4823$ & $	0.8502$ \\
         & & $Y_4$ & $8.6429$ & $	0.8770$ & $	0.8915$ \\
         & & $Y_5$ & $0.1597$ & $	0.2690$ & $	0.5696$ \\
         & & $Y_6$ & $3.7938$ & $	0.9266$ & $	0.9523$ \\    \cline{2-6}
           & \multirow{6}{*}{\makecell{TCN \\ model $1$}} & $Y_1$ & $0.0711$ & $	0.1376$ & $	0.8845$ \\
         & & $Y_2$ & $0.0021$ & $	0.0269$ & $	0.8440$ \\
         & & $Y_3$ & $0.4459$ & $	0.4904$ & $	0.8471$ \\
         & & $Y_4$ & $8.6066$ & $	0.8330$ & $	0.8919$ \\
         & & $Y_5$ & $0.1584$ & $	0.2762$ & $	0.5730$ \\
         & & $Y_6$ & $3.7958$ & $	0.9728$ & $	0.9523$ \\    \cline{2-6}
  & \multirow{6}{*}{\makecell{TCN \\ model $2$}} & $Y_1$ & $0.0711$ & $	0.1414$ & $	0.8844$ \\
         & & $Y_2$ & $0.0021$ & $	0.0262$ & $	0.8478$ \\
         & & $Y_3$ & $0.4334$ & $	0.4813$ & $	0.8514$ \\
         & & $Y_4$ & $8.7877$ & $	0.8595$ & $	0.8896$ \\
         & & $Y_5$ & $0.1516$ & $	0.2525$ & $	0.5914$ \\
         & & $Y_6$ & $3.8516$ & $	0.9888$ & $	0.9516$ \\     \hline
    \end{tabular}
\end{table}
\begin{table}[h] 
    \centering
    \caption{Performance metrics for PDE extraction phase for derivatives. Values in bold indicate the corresponding PDEs that are utilized.}\label{PDEextractionPhaseMetricsderivatives}
    \resizebox{\textwidth}{!}{ 
    \begin{tabular}{|c|c|c|c|c|c|}
        \hline
        Phase & Methods & Variable & MSE  & MAE & $R^2$ \\ \hline
\multirow{36}{*}{\makecell{PDE Extractions \\ $1958991$ training \\ $19000$ validation \\ $788$ testing}} & \multirow{6}{*}{SINDy} & $dY_1/ dX_1$ & $5.8788E^{+18}$ & $	2.2649E^{+09	}$ & $-2.7194E^{+16}$ \\
         & & $dY_2/dX_1$ & $7.09188E^{+17}$ & $	7.8661E^{+08	}$ & $-9.41825E^{+16}$ \\
         & & $dY_3/dX_1$ & $5.30668E^{+19}$ & $	6.8055E^{+09	}$ & $-3.32554E^{+16}$ \\
         & & $dY_4/dX_1$ & $2.71052E^{+19}$ & $	4.8655E^{+09	}$ & $-1.14348E^{+15}$ \\
         & & $dY_5/dX_1$ & $8.28482E^{+19}$ & $	8.5026E^{+09}$ & $	-1.61717E^{+17}$ \\
         & & $dY_6/dX_1$ & $8.98478E^{+18}$ & $	2.8000E^{+09}$ & $	-6.38548E^{+14}$ \\    \cline{2-6}
  & \multirow{6}{*}{LASSO} & $dY_1/ dX_1$  & $1.6382E^{+03}$ & $	1.6576E^{+01}$ & $	4.5876E^{-01}$ \\
         & & $dY_2/dX_1$ & $6.8290E^{+03}$ & $	3.1475E^{+01}$ & $	-2.4088E^{-01}$ \\
         & & $dY_3/dX_1$ & $2.1774E^{+03}$ & $	2.2091E^{+01}$ & $	1.6338E^{-01}$ \\
         & & $dY_4/dX_1$ & $6.3249E^{+03}$ & $	7.8828E^{+00}$ & $	3.4856E^{-02}$ \\
         & & $dY_5/dX_1$ & $1.5603E^{+02}$ & $	5.1895E^{+00}$ & $	-4.4799E^{-01}$ \\
         & & $dY_6/dX_1$ & $1.9095E^{+04}$ & $	5.0810E^{+01}$ & $	1.2204E^{-01}$ \\    \cline{2-6}
         & \multirow{6}{*}{B-LASSO} & $dY_1/ dX_1$ & $1.6237E^{+07}$ & $	2.1261E^{+02}$ & $	-4.6520E^{-01}$ \\
         & & $dY_2/dX_1$ & $3.9785E^{+07}$ & $	3.7705E^{+02}$ & $	-2.5150E^{-01}$ \\
         & & $dY_3/dX_1$ & $5.5873E^{+06}$ & $	1.8564E^{+02}$ & $	-1.9100E^{-02}$ \\
         & & $dY_4/dX_1$ & $1.5940E^{+13}$ & $	5.3401E^{+04}$ & $	-4.4738E^{+06}$ \\
         & & $dY_5/dX_1$ & $2.3453E^{+07}$ & $	1.2887E^{+02}$ & $	-2.5973E^{+00}$ \\
         & & $dY_6/dX_1$ & $4.2323E^{+12}$ & $	1.1728E^{+05}$ & $	-1.0066E^{+05}$ \\    \cline{2-6}
         & \multirow{6}{*}{CNN} & $dY_1/ dX_1$ & $0.0000$ & $	0.0000$ & $	0.9304$ \\
         & & $dY_2/dX_1$ & $0.0000$ & $	0.0000$ & $	0.4230$ \\
         & & $dY_3/dX_1$ & $0.0000$ & $	0.0001$ & $	0.5575$ \\
         & & $dY_4/dX_1$ & $0.0000$ & $	0.0000$ & $	0.9883$ \\
         & & $dY_5/dX_1$ & $0.0000$ & $	0.0000$ & $	0.9381$ \\
         & & $dY_6/dX_1$ & $0.0000$ & $	0.0002$ & $	0.8701$ \\    \cline{2-6}
  & \multirow{6}{*}{\makecell{TCN \\ model $1$ \\ vs. \\ model $2$}} & $dY_1/ dX_1$ & $0.0000\text{vs.}	0.0000$ & $	0.0000\text{vs.}	0.0000$ & $	0.8171\text{vs.}	\mathbf{0.9998}$ \\
         & & $dY_2/dX_1$ & $0.0002\text{vs.}	0.0000$ & $	0.0007\text{vs.}	0.0000$ & $	\mathbf{-0.4289}\text{vs.}	0.9312$ \\
         & & $dY_3/dX_1$ & $0.0000\text{vs.}	0.0000$ & $	0.0012\text{vs.}	0.0000$ & $	\mathbf{0.9168}\text{vs.}	0.8376$ \\
         & & $dY_4/dX_1$ & $0.0000\text{vs.}	0.0000$ & $	0.0001\text{vs.}	0.0000$ & $	0.5592\text{vs.}	0.2085$ \\
         & & $dY_5/dX_1$ & $0.0000\text{vs.}	0.0000$ & $	0.0012\text{vs.}	0.0000$ & $	0.7249\text{vs.}	0.9980$ \\
         & & $dY_6/dX_1$ & $0.0000\text{vs.}	0.0000$ & $	0.0008\text{vs.}	0.0000$ & $	\mathbf{0.9984}\text{vs.}	0.9922$ \\    \cline{2-6}
 & \multirow{6}{*}{SR} & $dY_1/ dX_1$ & $1.4906E^{+03}$ & $	1.9143E^{+01}$ & $	-4.5000E^{-03}$ \\
         & & $dY_2/dX_1$ & $1.5753E^{+03}$ & $	3.9556E^{+01}$ & $	-1.4758E^{+02}$ \\
         & & $dY_3/dX_1$ & $2.7606E^{+03}$ & $	3.1546E^{+01}$ & $	-2.2000E^{-03}$ \\
         & & $dY_4/dX_1$ & $2.8841E^{+05}$ & $	2.3984E^{+02}$ & $	-8.9000E^{-03}$ \\
         & & $dY_5/dX_1$ & $9.4075E^{+04}$ & $	1.2370E^{+02}$ & $	-1.2100E^{-02}$ \\
         & & $dY_6/dX_1$ & $1.2855E^{+04}$ & $	5.7653E^{+01}$ & $	-3.4000E^{-02}$ \\    \hline
    \end{tabular}
    } 
\end{table}
\clearpage
\begin{figure}[b] 
    \centering
    \begin{subfigure}[b]{0.95\textwidth} 
        \centering
        \includegraphics[width=\textwidth, height=8.6cm]{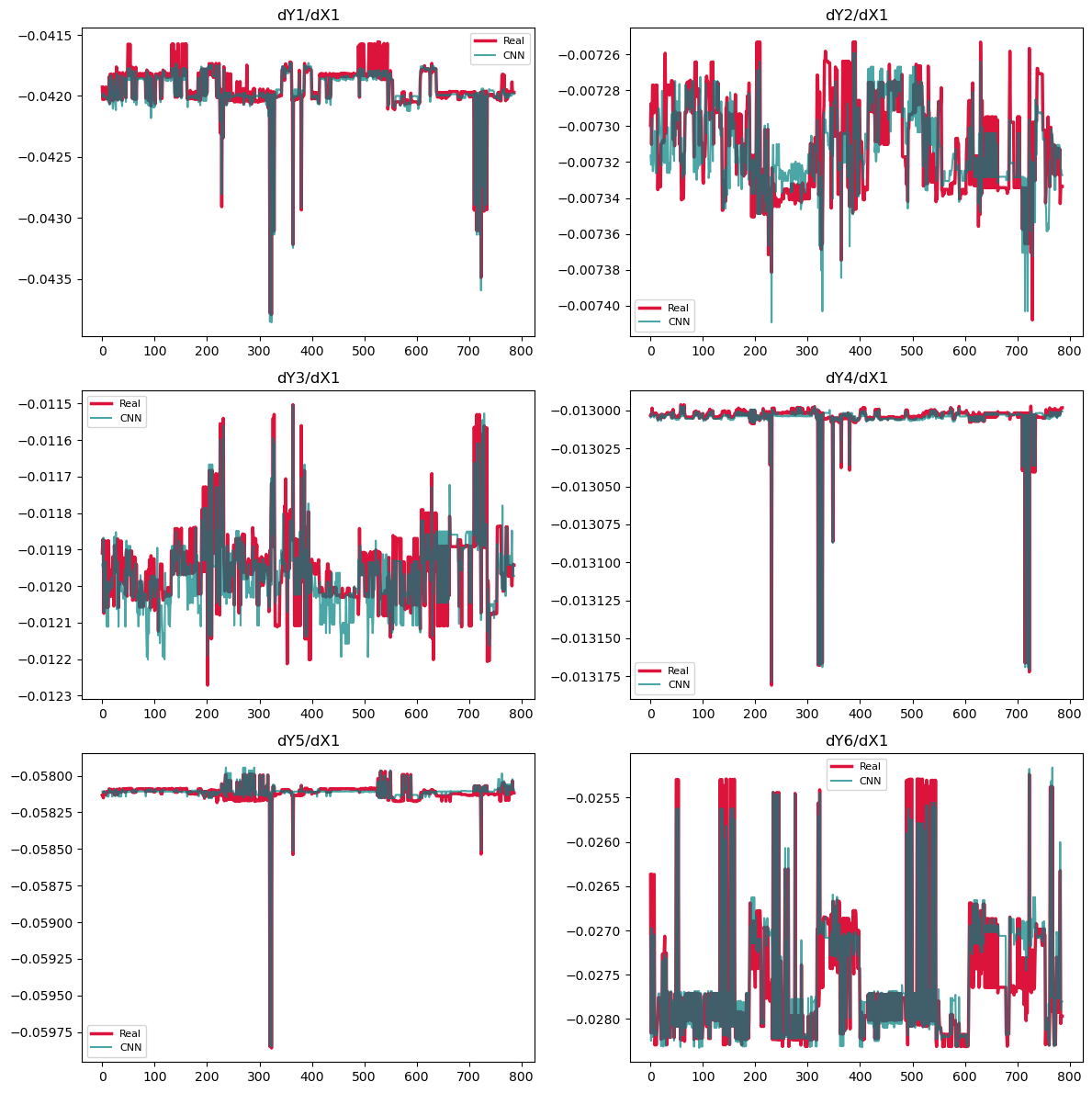} 
        \caption{CNN.}
        \label{DerivaPredCNN}
    \end{subfigure}
    \caption{Continues.} 
    \label{Predictions_PDEextraction_derivatives}
\end{figure}
\begin{figure}[h]
    \ContinuedFloat
    \centering
    \begin{subfigure}[b]{0.95\textwidth} 
        \centering
        \includegraphics[width=\textwidth, height=9cm]{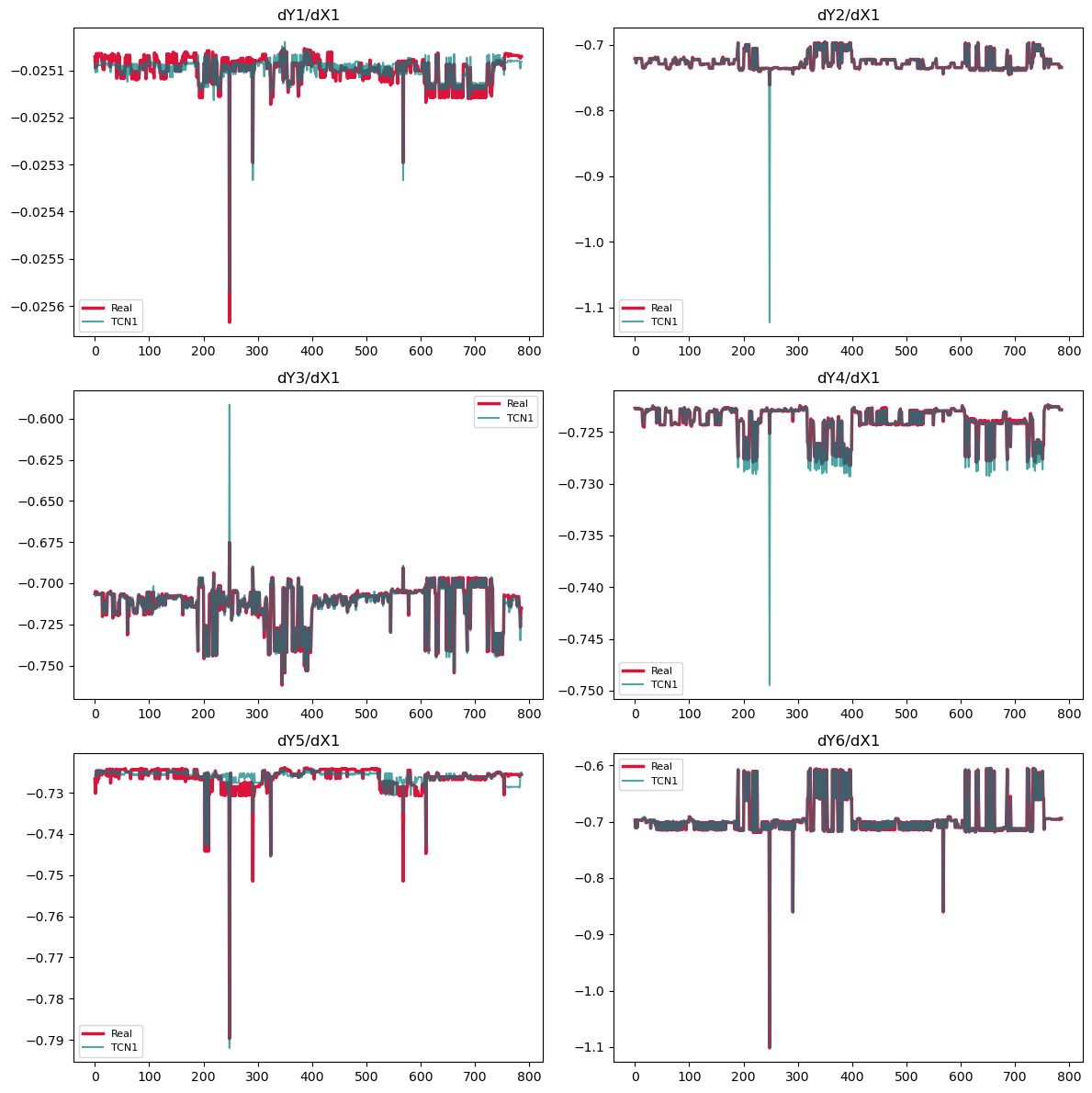} 
        \caption{TCN model $1$.}
        \label{DerivaPredTCN1}
    \end{subfigure}
    \vspace{0.5em} 
    \begin{subfigure}[b]{0.95\textwidth} 
        \centering
        \includegraphics[width=\textwidth, height=9cm]{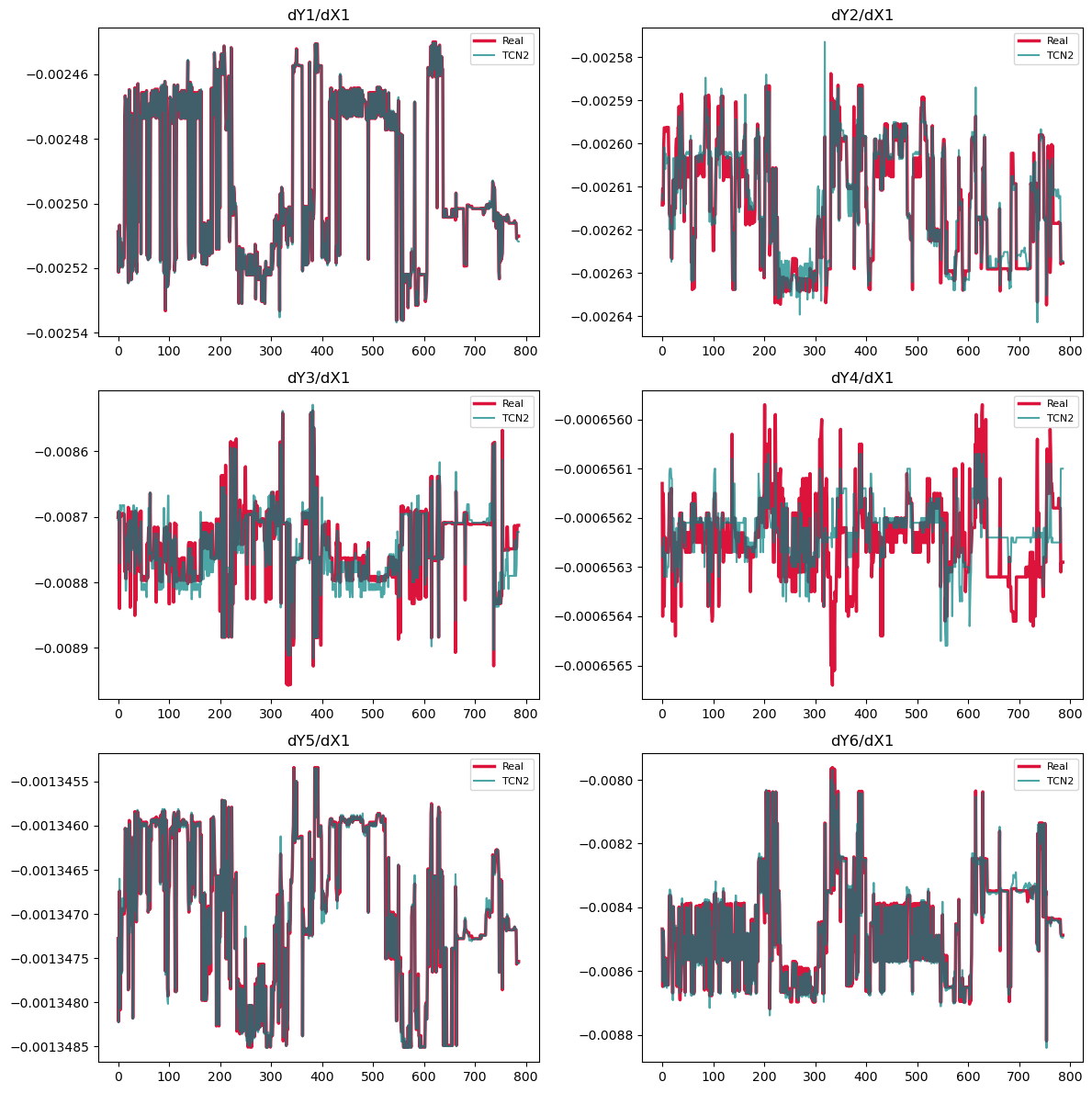} 
        \caption{TCN model $2$.}
        \label{DerivaPredTCN2}
    \end{subfigure}
    \caption{Predictions of the derivatives.} 
    \label{Predictions_PDEextraction}
\end{figure}
\begin{table}[h] 
    \centering
    \caption{Continues.}
    \begin{tabular}{|c|c|c|c|c|c|}
        \hline
        Phase & Methods & Variable & MSE &  MAE & $R^2$  \\ \hline
         \multirow{36}{*}{\makecell{Predictions \\ small samples}} & \multirow{6}{*}{\makecell{NN \\ $50$ Epochs}} & $Y_1$ &$0.1653$ & $	0.3158$ & $	\mathbf{0.3049}$ \\
         & & $Y_2$ & $0.0122	$ & $0.0924$ & $	0.3125$ \\
         & & $Y_3$ & $13.8082$ & $	3.3928$ & $	-2.7393$ \\
         & & $Y_4$ & $56.1014$ & $	3.7655$ & $	0.0000$ \\
         & & $Y_5$ & $0.1728$ & $	0.3347$ & $	0.4435$ \\
         & & $Y_6$ & $6.4717$ & $	1.8645	$ & $0.9132$ \\       \cline{2-6}
         & \multirow{6}{*}{\makecell{NN \\ Early-Stop}} & $Y_1$ & $0.2739$ & $	0.3846	$ & $-0.1520$ \\
         & & $Y_2$ & $0.0107$ & $	0.0777	$ & $0.3965$ \\
         & & $Y_3$ & $3.5552	$ & $1.4563	$ & $0.0373$ \\
         & & $Y_4$ & $36.5446	$ & $4.7508	$ & $0.0000$ \\
         & & $Y_5$ & $0.0984	$ & $0.1923	$ & $0.6829$ \\
         & & $Y_6$ & $5.2535	$ & $1.5618$ & $	0.9296$ \\         \cline{2-6}
         & \multirow{6}{*}{\makecell{PINN \\ $3$ PDEs \\ model $1$  \\ $50$ Epochs}} & $Y_1$ &  $0.4710$ & $	0.5466	$ & $-0.9808$ \\
         & & $Y_2$ & $0.0071$ & $	0.0552	$ & $0.5963$ \\
         & & $Y_3$ & $7.6201	$ & $2.1871$ & $	-1.0635$ \\
         & & $Y_4$ & $24.4053$ & $	4.0371$ & $	0.0000$ \\
         & & $Y_5$ & $0.2559	$ & $0.4373	$ & $0.1758$ \\
         & & $Y_6$ & $20.8505$ & $	3.7194	$ & $0.7205$ \\        \cline{2-6}
          & \multirow{6}{*}{\makecell{PINN \\ $3$ PDEs \\ model $1$ \\ Early-Stop}} & $Y_1$ & $0.4262	$ & $0.4646$ & $	-0.7922$ \\
         & & $Y_2$ & $0.0106	$ & $0.0641$ & $	0.3986$ \\
         & & $Y_3$ & $28.1717$ & $	4.8865$ & $	-6.6289$ \\
         & & $Y_4$ & $17.3826$ & $	2.9961	$ & $0.0000$ \\
         & & $Y_5$ & $0.1396	$ & $0.2520	$ & $0.5504$ \\
         & & $Y_6$ & $6.9809	$ & $1.9604$ & $	0.9064$ \\        \cline{2-6}
          & \multirow{6}{*}{\makecell{PINN \\ $1$ PDE \\ model $1$ \\ $50$ Epochs}} & $Y_1$ & $0.2242$ & $0.3700$ & $0.0570$ \\
         & & $Y_2$ & $0.0067$ & $0.0590	$ & $0.6195$ \\
         & & $Y_3$ & $5.8388$ &  $1.7115	$ & $-0.5811$ \\
         & & $Y_4$ & $21.3012$ & $	3.1901$ & $	0.0000$ \\
         & & $Y_5$ & $0.0834	$ & $0.1867$ & $	\mathbf{0.7313}$ \\
         & & $Y_6$ & $9.6882	$ & $2.3592	$ & $0.8701$ \\        \cline{2-6}
          & \multirow{6}{*}{\makecell{PINN \\ $1$ PDE \\ model $1$ \\ Early-Stop}} & $Y_1$ & $1.5982	$ & $1.0737	$ & $-5.7206$ \\
         & & $Y_2$ & $0.0122	$ & $0.0728	$ & $0.3071$ \\
         & & $Y_3$ & $18.1349	$ & $2.6480	$ & $-3.9109$ \\
         & & $Y_4$ & $33.7818	$ & $4.1057	$ & $0.0000$ \\
         & & $Y_5$ & $0.4207	$ & $0.5274	$ & $-0.3551$ \\
         & & $Y_6$ & $9.7196	$ & $2.2411	$ & $0.8697$ \\         \hline
    \end{tabular}
\end{table}
\begin{table}[h] 
    \ContinuedFloat
    \centering
    \caption{Continues.}
    \begin{tabular}{|c|c|c|c|c|c|}
        \hline
        Phase & Methods & Variable & MSE &  MAE & $R^2$  \\ \hline
         \multirow{36}{*}{\makecell{Predictions \\ small samples}} & \multirow{6}{*}{\makecell{PINN \\ $1$ PDE \\ model $2$ \\ $50$ Epochs}} & $Y_1$ & $0.2589$ & $	0.4134$ & $	-0.0888$ \\
         & & $Y_2$ & $0.0075$ & $	0.0560$ & $	0.5733$ \\
         & & $Y_3$ & $6.4928$ & $	2.0941$ & $	-0.7583$ \\
         & & $Y_4$ & $7.8856$ & $	2.2620$ & $	0.0000$ \\
         & & $Y_5$ & $0.1046$ & $	0.1953$ & $	0.6632$ \\
         & & $Y_6$ & $7.4993$ & $	2.2371$ & $	0.8995$ \\        \cline{2-6}
          & \multirow{6}{*}{\makecell{PINN \\ $1$ PDE \\ model $2$ \\ Early-Stop}} & $Y_1$ &  $0.4174$ & $	0.5433	$ & $-0.7550$ \\
         & & $Y_2$ & $0.0065$ & $	0.0603	$ & $\mathbf{0.6327}$ \\
         & & $Y_3$ & $1.5103$ & $	0.9743	$ & $\mathbf{0.5910}$ \\
         & & $Y_4$ & $8.4065$ & $	2.3679	$ & $0.0000$ \\
         & & $Y_5$ & $0.1501$ & $	0.2841	$ & $0.5164$ \\
         & & $Y_6$ & $2.8607$ & $	1.0207	$ & $\mathbf{0.9616}$ \\         \cline{2-6}
          & \multirow{6}{*}{\makecell{BLR \\ $230$ samples \\ $200$ tuning \\ $1$ chain}} & $Y_1$ & $0.3252$ & $	0.4378$ & $	-0.3674$ \\
         & & $Y_2$ & $0.0192$ & $	0.0846$ & $	-0.0887$ \\
         & & $Y_3$ & $4.4595$ & $	1.6765$ & $	-0.2076^*$ \\
         & & $Y_4$ & $2.9796$ & $	1.4308	$ & $0.0000$ \\
         & & $Y_5$ & $0.3713	$ & $0.4472$ & $	-0.1958$ \\
         & & $Y_6$ & $53.1858$ & $	6.1492$ & $	0.2869$ \\         \cline{2-6}
          & \multirow{6}{*}{\makecell{PI-BLR \\ $3$ PDEs \\ model $1$ \\ $230$ samples \\ $200$ tuning \\ $1$ chain}} & $Y_1$ & $0.3319$ & $	0.4402	$ & $-0.3957$ \\
         & & $Y_2$ & $0.0195	$ & $0.0840	$ & $-0.1033$ \\
         & & $Y_3$ & $4.5982	$ & $1.7016	$ & $-0.2452$ \\
         & & $Y_4$ & $2.5773	$ & $1.3180	$ & $0.0000$ \\
         & & $Y_5$ & $0.3632	$ & $0.4460	$ & $-0.1698$ \\
         & & $Y_6$ & $48.1624$ & $	5.7915	$ & $0.3543$ \\         \cline{2-6}
          & \multirow{6}{*}{\makecell{BNN \\ $10$ neurons \\ $2$ hidden layers \\ $230$ samples \\ $200$ tuning \\ $1$ chain}} & $Y_1$ & $0.3586$ & $	0.4752	$ & $-0.5079$ \\
         & & $Y_2$ & $0.0183	$ & $0.0798$ & $	-0.0334$ \\
         & & $Y_3$ & $8.2709	$ & $2.3538$ & $	-1.2398$ \\
         & & $Y_4$ & $0.8176	$ & $0.8727$ & $	0.0000$ \\
         & & $Y_5$ & $0.3127	$ & $0.4534$ & $	-0.0072^*$ \\
         & & $Y_6$ & $79.5339	$ & $8.1894$ & $	-0.0663$ \\        \cline{2-6}
          & \multirow{6}{*}{\makecell{B-PINN \\ $3$ PDEs, model $1$ \\ $10$ neurons \\ $2$ hidden layers \\ $230$ samples \\ $200$ tuning, $1$ chain}} & $Y_1$ & $0.3003$ & $	0.4290$ & $	-0.2218^*$ \\
         & & $Y_2$ & $0.0184	$ & $0.0922$ & $	-0.0138^*$ \\
         & & $Y_3$ & $9.9477	$ & $2.6143$ & $	-1.6404$ \\
         & & $Y_4$ & $1.2192	$ & $1.0707$ & $	0.0000$ \\
         & & $Y_5$ & $1.1319	$ & $0.9782$ & $	-2.5539$ \\
         & & $Y_6$ & $73.3609	$ & $8.3759$ & $	0.0012^*$ \\       \hline
    \end{tabular}
\end{table}
\begin{table}[h] 
    \ContinuedFloat
    \centering
    \caption{Continues.}
    \begin{tabular}{|c|c|c|c|c|c|}
        \hline
        Phase & Methods & Variable & MSE &  MAE & $R^2$  \\ \hline
\multirow{36}{*}{\makecell{Predictions \\ medium samples}} & \multirow{6}{*}{\makecell{NN \\ $50$ Epochs}} & $Y_1$ & $0.0493$ & $	0.1353	$ & $\mathbf{0.7925}$ \\
         & & $Y_2$ & $0.0056$ & $	0.0463	$ & $0.6845$ \\
         & & $Y_3$ & $0.8598	$ & $0.7093$ & $	0.7672$ \\
         & & $Y_4$ & $1.3986	$ & $0.6100	$ & $0.0000$ \\
         & & $Y_5$ & $0.5047	$ & $0.5316	$ & $-0.6255$ \\
         & & $Y_6$ & $1.7182	$ & $0.6775	$ & $0.9770$ \\         \cline{2-6}
         & \multirow{6}{*}{\makecell{NN \\ Early-Stop}} & $Y_1$ & $0.0762$ & $	0.1651$ & $	0.6798$ \\
         & & $Y_2$ & $0.0046$ & $	0.0372$ & $	0.7391$ \\
         & & $Y_3$ & $0.9166$ & $	0.7697$ & $	0.7518$ \\
         & & $Y_4$ & $1.1903	$ & $0.6661	$ & $0.0000$ \\
         & & $Y_5$ & $0.4131$ & $	0.5076	$ & $-0.3305$ \\
         & & $Y_6$ & $1.6525	$ & $0.6295$ & $	0.9778$ \\        \cline{2-6}
         & \multirow{6}{*}{\makecell{PINN \\ $3$ PDEs \\ model $1$  \\ $50$ Epochs}} & $Y_1$ & $0.0706	$ & $0.1731	$ & $0.7031$ \\
         & & $Y_2$ & $0.0054	$ & $0.0462	$ & $0.6957$ \\
         & & $Y_3$ & $0.6627	$ & $0.6286	$ & $0.8205$ \\
         & & $Y_4$ & $0.6099	$ & $0.5457	$ & $0.0000$ \\
         & & $Y_5$ & $0.6838	$ & $0.6520	$ & $-1.2024$ \\
         & & $Y_6$ & $2.1924	$ & $0.7246	$ & $0.9706$ \\         \cline{2-6}
          & \multirow{6}{*}{\makecell{PINN \\ $3$ PDEs \\ model $1$ \\ Early-Stop}} & $Y_1$ & $0.0549$ & $	0.1535	$ & $0.7693$ \\
         & & $Y_2$ & $0.0041$ & $	0.0369	$ & $0.7679$ \\
         & & $Y_3$ & $0.5982$ & $	0.5925	$ & $0.8380$ \\
         & & $Y_4$ & $1.7424$ & $	0.9240	$ & $0.0000$ \\
         & & $Y_5$ & $1.3734$ & $	0.8931	$ & $-3.4236$ \\
         & & $Y_6$ & $1.8844	$ & $0.7179$ & $	0.9747$ \\       \cline{2-6}
      & \multirow{6}{*}{\makecell{PINN \\ $1$ PDE \\ model $1$ \\ $50$ Epochs}} & $Y_1$ & $0.0972$ & $	0.2356$ & $	0.5914$ \\
         & & $Y_2$ & $0.0050$ & $	0.0508$ & $	0.7145$ \\
         & & $Y_3$ & $1.0653$ & $	0.8207$ & $	0.7115$ \\
         & & $Y_4$ & $1.2751$ & $	0.6275$ & $	0.0000$ \\
         & & $Y_5$ & $0.3296$ & $	0.3793$ & $	-0.0617$ \\
         & & $Y_6$ & $1.9017$ & $	0.6159$ & $	0.9745$ \\        \cline{2-6}
          & \multirow{6}{*}{\makecell{PINN \\ $1$ PDE \\ model $1$ \\ Early-Stop}} & $Y_1$ &  $0.0534	$ & $0.1545$ & $	0.7756$ \\
         & & $Y_2$ & $0.0038	$ & $0.0366$ & $	0.7840$ \\
         & & $Y_3$ & $0.5446	$ & $0.5817$ & $	0.8525$ \\
         & & $Y_4$ & $1.4381	$ & $0.8962$ & $	0.0000$ \\
         & & $Y_5$ & $0.7744	$ & $0.7441$ & $	-1.4944$ \\
         & & $Y_6$ & $1.4532	$ & $0.6028$ & $	0.9805$ \\         \hline
    \end{tabular}
\end{table}
\begin{table}[h] 
    \ContinuedFloat
    \centering
    \caption{Continues.}
    \begin{tabular}{|c|c|c|c|c|c|}
        \hline
        Phase & Methods & Variable & MSE &  MAE & $R^2$  \\ \hline
         \multirow{36}{*}{\makecell{Predictions \\ medium samples}} & \multirow{6}{*}{\makecell{PINN \\ $1$ PDE \\ model $2$ \\ $50$ Epochs}} & $Y_1$ & $0.0519$ & $	0.1485$ & $	0.7816$ \\
         & & $Y_2$ & $0.0041$ & $	0.0342$ & $	0.7708$ \\
         & & $Y_3$ & $0.5349$ & $	0.5641$ & $	\mathbf{0.8552}$ \\
         & & $Y_4$ & $0.5675$ & $	0.5921$ & $	0.0000$ \\
         & & $Y_5$ & $0.2466$ & $	0.3829$ & $	\mathbf{0.2056}$ \\
         & & $Y_6$ & $1.3562$ & $	0.4935$ & $	\mathbf{0.9818}$ \\         \cline{2-6}
          & \multirow{6}{*}{\makecell{PINN \\ $1$ PDE \\ model $2$ \\ Early-Stop}} & $Y_1$ & $0.0530$ & $	0.1393$ & $	0.7770$ \\
         & & $Y_2$ & $0.0038$ & $	0.0354$ & $	\mathbf{0.7849}$ \\
         & & $Y_3$ & $0.9487$ & $	0.7718$ & $	0.7431$ \\
         & & $Y_4$ & $0.7525$ & $	0.5272$ & $	0.0000$ \\
         & & $Y_5$ & $1.4244$ & $	1.0116$ & $	-3.5878$ \\
         & & $Y_6$ & $1.5919$ & $	0.5432$ & $	0.9787$ \\         \cline{2-6}
   & \multirow{6}{*}{\makecell{BLR \\ $230$ samples \\ $200$ tuning \\ $1$ chain}} & $Y_1$ & $0.3061$ & $	0.4203$ & $	-0.2872$ \\
         & & $Y_2$ & $0.0199$ & $	0.0857$ & $	-0.1250$ \\
         & & $Y_3$ & $4.5271$ & $	1.6927$ & $	-0.2259$ \\
         & & $Y_4$ & $3.5336$ & $	1.7652$ & $	0.0000$ \\
         & & $Y_5$ & $1.3282$ & $	0.9886$ & $	-3.2779$ \\
         & & $Y_6$ & $45.0573$ & $	5.3436$ & $	0.3959$ \\         \cline{2-6}
     & \multirow{6}{*}{\makecell{PI-BLR \\ $3$ PDEs \\ model $1$ \\ $230$ samples \\ $200$ tuning \\ $1$ chain}} & $Y_1$ & $0.3136$ & $	0.4240$ & $	-0.3186$ \\
         & & $Y_2$ & $0.0198$ & $	0.0863$ & $	-0.1193$ \\
         & & $Y_3$ & $4.4868$ & $	1.6876$ & $	-0.2150^*$ \\
         & & $Y_4$ & $2.5198$ & $	1.4567$ & $	0.0000$ \\
         & & $Y_5$ & $1.3857$ & $	1.0130$ & $	-3.4632$ \\
         & & $Y_6$ & $44.7380$ & $	5.2786$ & $	0.4002$ \\         \cline{2-6}
       & \multirow{6}{*}{\makecell{BNN \\ $10$ neurons \\ $2$ hidden layers \\ $230$ samples \\ $200$ tuning \\ $1$ chain}} & $Y_1$ & $0.2693$ & $	0.3981$ & $	-0.1326$ \\
         & & $Y_2$ & $0.0177$ & $	0.0836$ & $	0.0006^*$ \\
         & & $Y_3$ & $4.7358$ & $	1.7381$ & $	-0.2825$ \\
         & & $Y_4$ & $1.7934	$ & $1.3088$ & $	0.0000$ \\
         & & $Y_5$ & $0.8035$ & $	0.8207	$ & $-1.5880$ \\
         & & $Y_6$ & $75.4659$ & $	8.4365$ & $	-0.0118$ \\        \cline{2-6}
    & \multirow{6}{*}{\makecell{B-PINN \\ $3$ PDEs, model $1$ \\ $10$ neurons \\ $2$ hidden layers \\ $230$ samples \\ $200$ tuning, $1$ chain}} & $Y_1$ &$0.2675$ & $	0.3989$ & $	-0.1248^*$ \\
         & & $Y_2$ & $0.0178$ & $	0.0841$ & $	-0.0043$ \\
         & & $Y_3$ & $4.8732$ & $	1.7617$ & $	-0.3197$ \\
         & & $Y_4$ & $1.7755$ & $	1.3020$ & $	0.0000$ \\
         & & $Y_5$ & $0.7827$ & $	0.8106$ & $	-1.5210^*$ \\
         & & $Y_6$ & $74.7314$ & $	8.3907$ & $	-0.0019^*$ \\        \hline
    \end{tabular}
\end{table}
\begin{table}[h] 
    \ContinuedFloat
    \centering
    \caption{Continues.}
    \begin{tabular}{|c|c|c|c|c|c|}
        \hline
        Phase & Methods & Variable & MSE &  MAE & $R^2$  \\ \hline
         \multirow{36}{*}{\makecell{Predictions \\ large samples}} & \multirow{6}{*}{\makecell{NN \\ $50$ Epochs}} & $Y_1$ & $0.0379$ & $	0.1157$ & $	0.8459$ \\
         & & $Y_2$ & $0.0043$ & $	0.0351$ & $	0.7655$ \\
         & & $Y_3$ & $0.6136$ & $	0.6095$ & $	0.8371$ \\
         & & $Y_4$ & $0.4036$ & $	0.6082$ & $	0.0000$ \\
         & & $Y_5$ & $0.6496$ & $	0.6310$ & $	-1.0394$ \\
         & & $Y_6$ & $1.4037	$ & $0.7219$ & $	0.9809$ \\         \cline{2-6}
         & \multirow{6}{*}{\makecell{NN \\ Early-Stop}} & $Y_1$ & $0.0367$ & $	0.1044$ & $	\mathbf{0.8506}$ \\
         & & $Y_2$ & $0.0041$ & $	0.0317$ & $	0.7749$ \\
         & & $Y_3$ & $0.6614$ & $	0.6423$ & $	0.8244$ \\
         & & $Y_4$ & $0.0658$ & $	0.1812$ & $	0.0000$ \\
         & & $Y_5$ & $0.2918$ & $	0.3825$ & $	0.0838$ \\
         & & $Y_6$ & $0.9804$ & $	0.3697$ & $	0.9867$ \\        \cline{2-6}
   & \multirow{6}{*}{\makecell{PINN \\ $3$ PDEs \\ model $1$  \\ $50$ Epochs}} & $Y_1$ & $0.0578$ & $	0.1628$ & $	0.7649$ \\
         & & $Y_2$ & $0.0040$ & $	0.0325$ & $	0.7784$ \\
         & & $Y_3$ & $0.5438$ & $	0.5614$ & $	0.8557$ \\
         & & $Y_4$ & $0.0778$ & $	0.1975$ & $	0.0000$ \\
         & & $Y_5$ & $0.2592$ & $	0.3673$ & $	0.1863$ \\
         & & $Y_6$ & $1.3535$ & $	0.3338$ & $	0.9816$ \\        \cline{2-6}
    & \multirow{6}{*}{\makecell{PINN \\ $3$ PDEs \\ model $1$ \\ Early-Stop}} & $Y_1$ & $0.0419$ & $	0.1221$ & $	0.8293$ \\
         & & $Y_2$ & $0.0040$ & $	0.0325$ & $	0.7795$ \\
         & & $Y_3$ & $0.5847$ & $	0.5921$ & $	0.8448$ \\
         & & $Y_4$ & $0.1018$ & $	0.2574$ & $	0.0000$ \\
         & & $Y_5$ & $0.2672$ & $	0.3596$ & $	0.1612$ \\
         & & $Y_6$ & $0.9316$ & $	0.3786$ & $	0.9873$ \\        \cline{2-6}
       & \multirow{6}{*}{\makecell{PINN \\ $1$ PDE \\ model $1$ \\ $50$ Epochs}} & $Y_1$ & $0.0457$ & $	0.1328$ & $	0.8142$ \\
         & & $Y_2$ & $0.0040$ & $	0.0331$ & $	0.7778$ \\
         & & $Y_3$ & $0.6872$ & $	0.6369$ & $	0.8176$ \\
         & & $Y_4$ & $0.1530$ & $	0.3383$ & $	0.0000$ \\
         & & $Y_5$ & $0.1562$ & $	0.2529$ & $	\mathbf{0.5096}$ \\
         & & $Y_6$ & $1.3277$ & $	0.3246$ & $	0.9819$ \\        \cline{2-6}
   & \multirow{6}{*}{\makecell{PINN \\ $1$ PDE \\ model $1$ \\ Early-Stop}} & $Y_1$ & $0.0534$ & $	0.1547$ & $	0.7827$ \\
         & & $Y_2$ & $0.0039$ & $	0.0310$ & $	\mathbf{0.7877}$ \\
         & & $Y_3$ & $0.6382$ & $	0.6238$ & $	0.8306$ \\
         & & $Y_4$ & $0.0573$ & $	0.1453$ & $	0.0000$ \\
         & & $Y_5$ & $0.1945$ & $	0.2983$ & $	0.3894$ \\
         & & $Y_6$ & $1.0142$ & $	0.2869$ & $	0.9862$ \\         \hline
    \end{tabular}
\end{table}
\begin{table}[h] 
    \ContinuedFloat
    \centering
    \caption{Performance metrics for algorithms in predictions: Bold values show the best predictions for each sample size; star values denote the top Bayesian predictions.}\label{PredictionPhaseMetrics}
    \begin{tabular}{|c|c|c|c|c|c|}
        \hline
        Phase & Methods & Variable & MSE &  MAE & $R^2$  \\ \hline
         \multirow{36}{*}{\makecell{Predictions \\ large samples}} & \multirow{6}{*}{\makecell{PINN \\ $1$ PDE \\ model $2$ \\ $50$ Epochs}} & $Y_1$ & $0.0437$ & $	0.1061$ & $	0.8223$ \\
         & & $Y_2$ & $0.0041$ & $	0.0343$ & $	0.7759$ \\
         & & $Y_3$ & $0.4906$ & $	0.5483$ & $	\mathbf{0.8698}$ \\
         & & $Y_4$ & $0.0524$ & $	0.1625$ & $	0.0000$ \\
         & & $Y_5$ & $0.3354$ & $	0.4038$ & $	-0.0531$ \\
         & & $Y_6$ & $0.7349$ & $	0.3176$ & $	\mathbf{0.9900}$ \\         \cline{2-6}
   & \multirow{6}{*}{\makecell{PINN \\ $1$ PDE \\ model $2$ \\ Early-Stop}} & $Y_1$ & $0.0671$ & $	0.1869$ & $	0.7270$ \\
         & & $Y_2$ & $0.0039$ & $	0.0324$ & $	0.7842$ \\
         & & $Y_3$ & $0.6033$ & $	0.5940$ & $	0.8399$ \\
         & & $Y_4$ & $0.2176$ & $	0.2832$ & $	0.0000$ \\
         & & $Y_5$ & $0.6023$ & $	0.6148$ & $	-0.8910$ \\
         & & $Y_6$ & $1.3005$ & $	0.4777$ & $	0.9823$ \\         \cline{2-6}
       & \multirow{6}{*}{\makecell{BLR \\ $230$ samples \\ $200$ tuning \\ $1$ chain}} & $Y_1$ & $0.3052$ & $	0.4228$ & $	-0.2417$ \\
         & & $Y_2$ & $0.0215$ & $	0.0916$ & $	-0.1795$ \\
         & & $Y_3$ & $4.5046$ & $	1.6909$ & $	-0.1957$ \\
         & & $Y_4$ & $2.7333$ & $	1.5126$ & $	0.0000$ \\
         & & $Y_5$ & $1.5531$ & $	1.0503$ & $	-3.8761$ \\
         & & $Y_6$ & $46.8142$ & $	5.5291$ & $	0.3626$ \\         \cline{2-6}
      & \multirow{6}{*}{\makecell{PI-BLR \\ $3$ PDEs \\ model $1$ \\ $230$ samples \\ $200$ tuning \\ $1$ chain}} & $Y_1$ & $0.3074$ & $	0.4258$ & $	-0.2506$ \\
         & & $Y_2$ & $0.0216$ & $	0.0909$ & $	-0.1859$ \\
         & & $Y_3$ & $4.4982$ & $	1.6922$ & $	-0.1940$ \\
         & & $Y_4$ & $1.9658$ & $	1.2354$ & $	0.0000$ \\
         & & $Y_5$ & $1.6085$ & $	1.0786$ & $	-4.0502$ \\
         & & $Y_6$ & $46.7515$ & $	5.4492$ & $	0.3635$ \\         \cline{2-6}
          & \multirow{6}{*}{\makecell{BNN \\ $10$ neurons \\ $2$ hidden layers \\ $230$ samples \\ $200$ tuning \\ $1$ chain}} & $Y_1$ & $0.3045$ & $	0.4331$ & $	-0.2390$ \\
         & & $Y_2$ & $0.0185$ & $	0.0922$ & $	-0.0171$ \\
         & & $Y_3$ & $9.9584$ & $	2.6197$ & $	-1.6433$ \\
         & & $Y_4$ & $1.2116$ & $	1.0683$ & $	0.0000$ \\
         & & $Y_5$ & $1.1297$ & $	0.9758$ & $	-2.5469^*$ \\
         & & $Y_6$ & $73.6081$ & $	8.3863$ & $	-0.0021$ \\         \cline{2-6}
          & \multirow{6}{*}{\makecell{B-PINN \\ $3$ PDEs, model $1$ \\ $10$ neurons \\ $2$ hidden layers \\ $230$ samples \\ $200$ tuning, $1$ chain}} & $Y_1$ & $0.3003$ & $	0.4290$ & $	-0.2218^*$ \\
         & & $Y_2$ & $0.0184$ & $	0.0922$ & $	-0.0138^*$ \\
         & & $Y_3$ & $9.9477$ & $	2.6143$ & $	-1.6404^*$ \\
         & & $Y_4$ & $1.2192$ & $	1.0707$ & $	0.0000$ \\
         & & $Y_5$ & $1.1319$ & $	0.9782$ & $	-2.5539$ \\
         & & $Y_6$ & $73.3609$ & $	8.3759$ & $	0.0012^*$ \\       \hline
    \end{tabular}
\end{table}
\begin{figure}[h] 
    \centering
    \begin{subfigure}[b]{0.95\textwidth} 
        \centering
        \includegraphics[width=\textwidth, height=9cm]{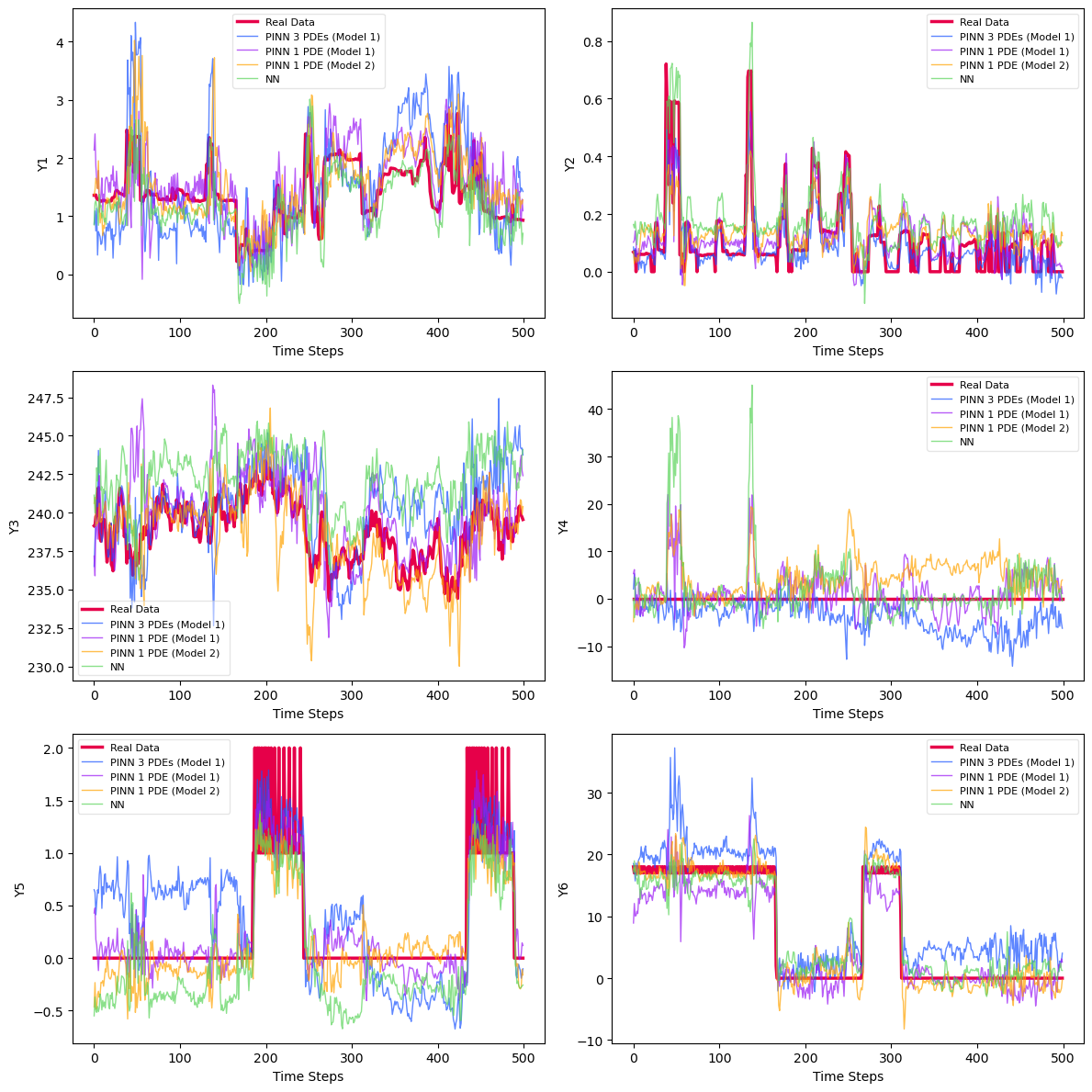} 
        \caption{Predictions of the small samples with $50$ epochs.}
        \label{PINN_50Epochs_2000}
    \end{subfigure}
    \vspace{0.5em} 
    \begin{subfigure}[b]{0.95\textwidth} 
        \centering
        \includegraphics[width=\textwidth, height=9cm]{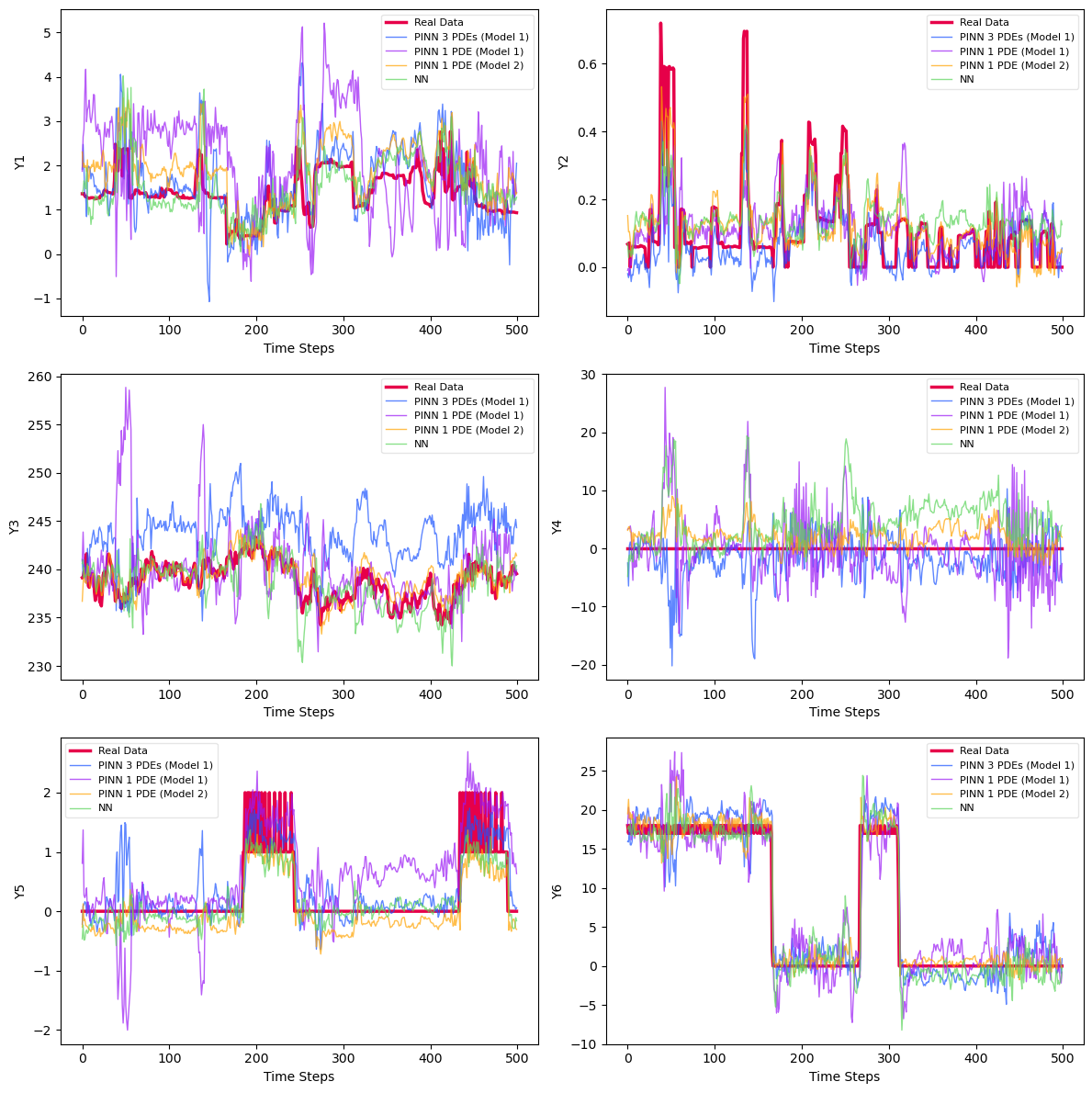} 
        \caption{Predictions of the small samples with early stopping.}
        \label{PINN_EarlyStop_2000}
    \end{subfigure}
    \caption{Continues.} 
    \label{Plotsss}
\end{figure}
\begin{figure}[h]
    \ContinuedFloat
    \centering
    \begin{subfigure}[b]{0.95\textwidth} 
        \centering
        \includegraphics[width=\textwidth, height=9cm]{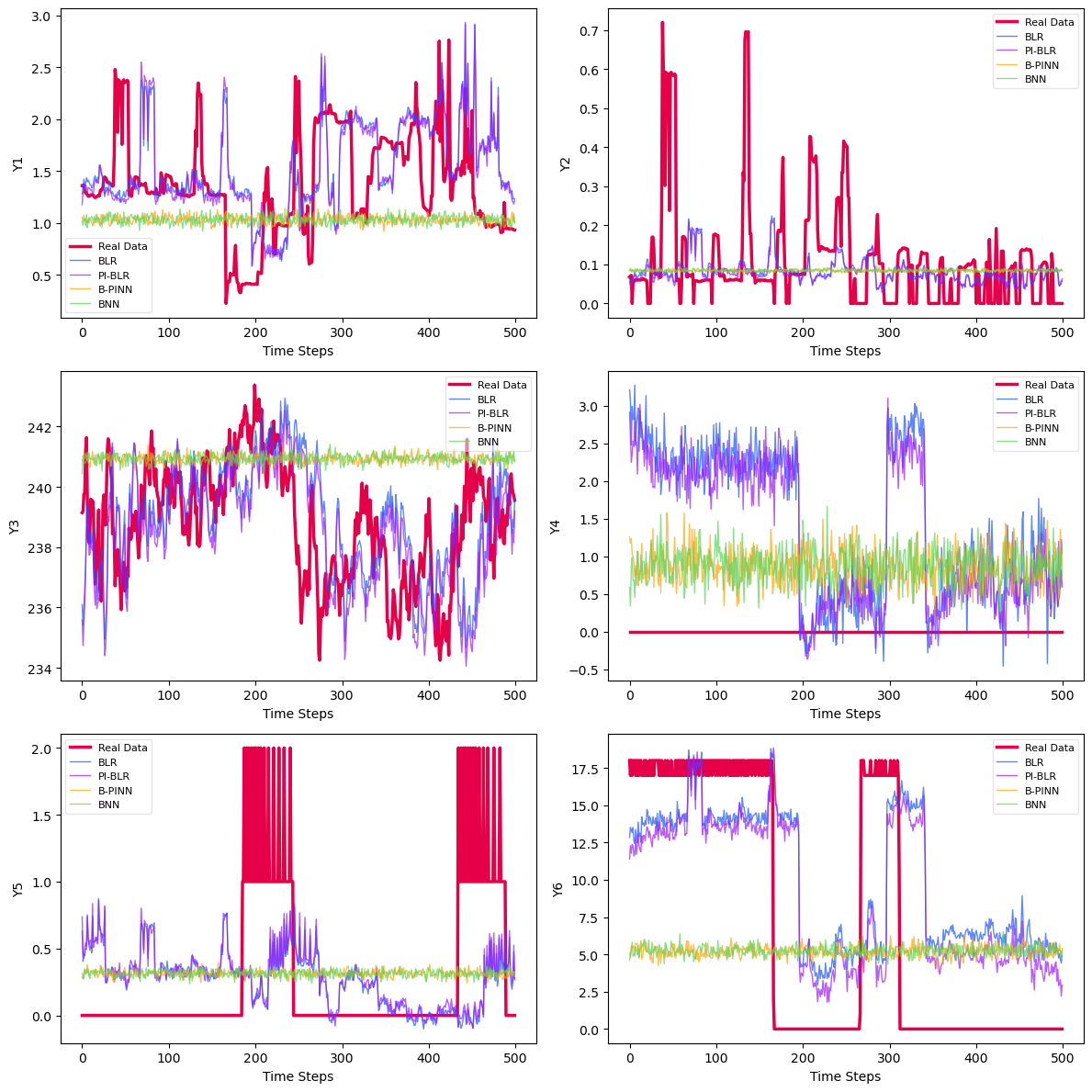} 
        \caption{Bayesian predictions of the small samples.}
        \label{Bayesian_2000}
    \end{subfigure}
    \vspace{0.5em} 
    \begin{subfigure}[b]{0.95\textwidth} 
        \centering
        \includegraphics[width=\textwidth, height=9cm]{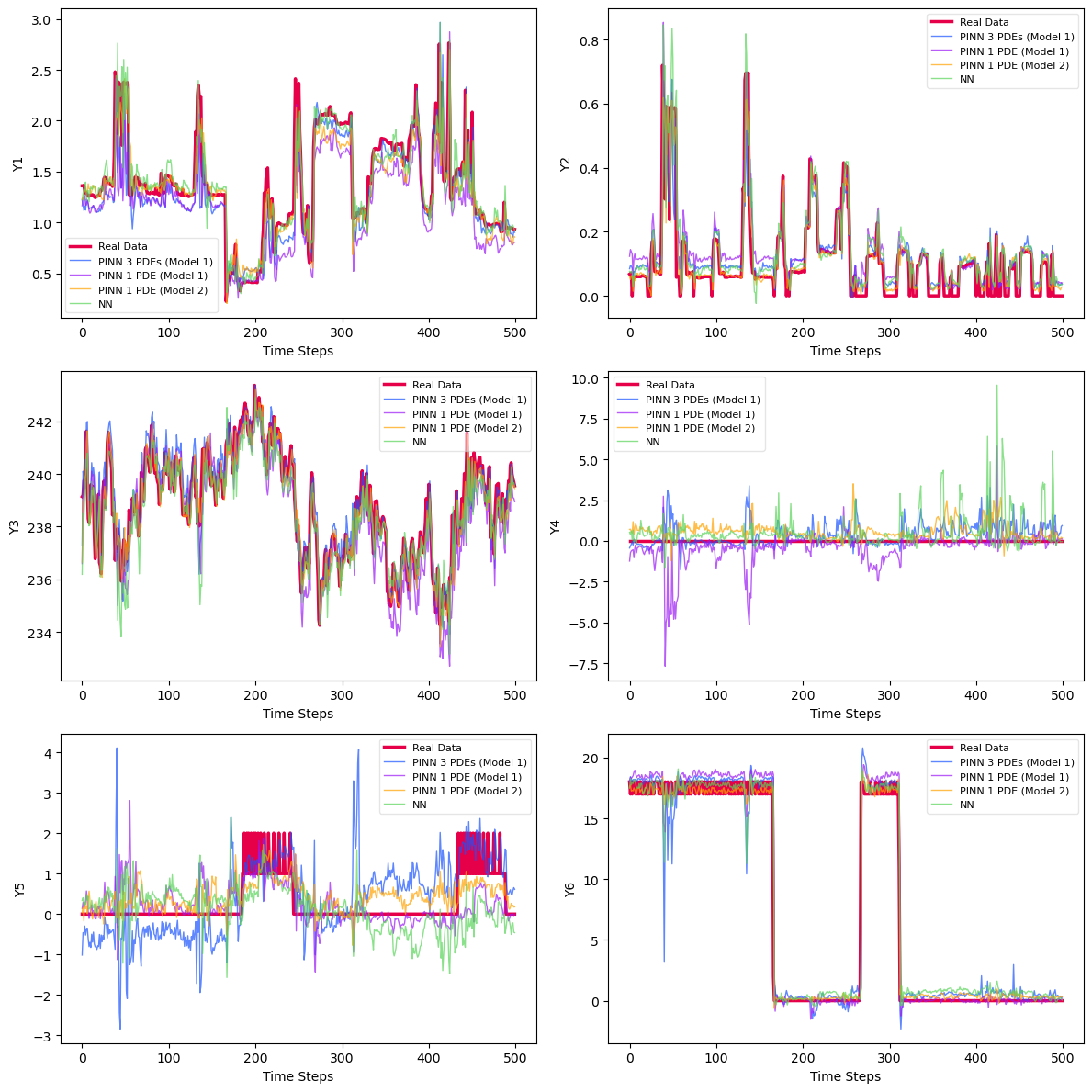} 
        \caption{Predictions of the medium-size samples with $50$ epochs.}
        \label{PINN_50Epochs_26000}
    \end{subfigure}
    \caption{Continues.} 
    \label{Plotsss}
\end{figure}
\begin{figure}[h]
    \ContinuedFloat
    \centering
    \begin{subfigure}[b]{0.95\textwidth} 
        \centering
        \includegraphics[width=\textwidth, height=9cm]{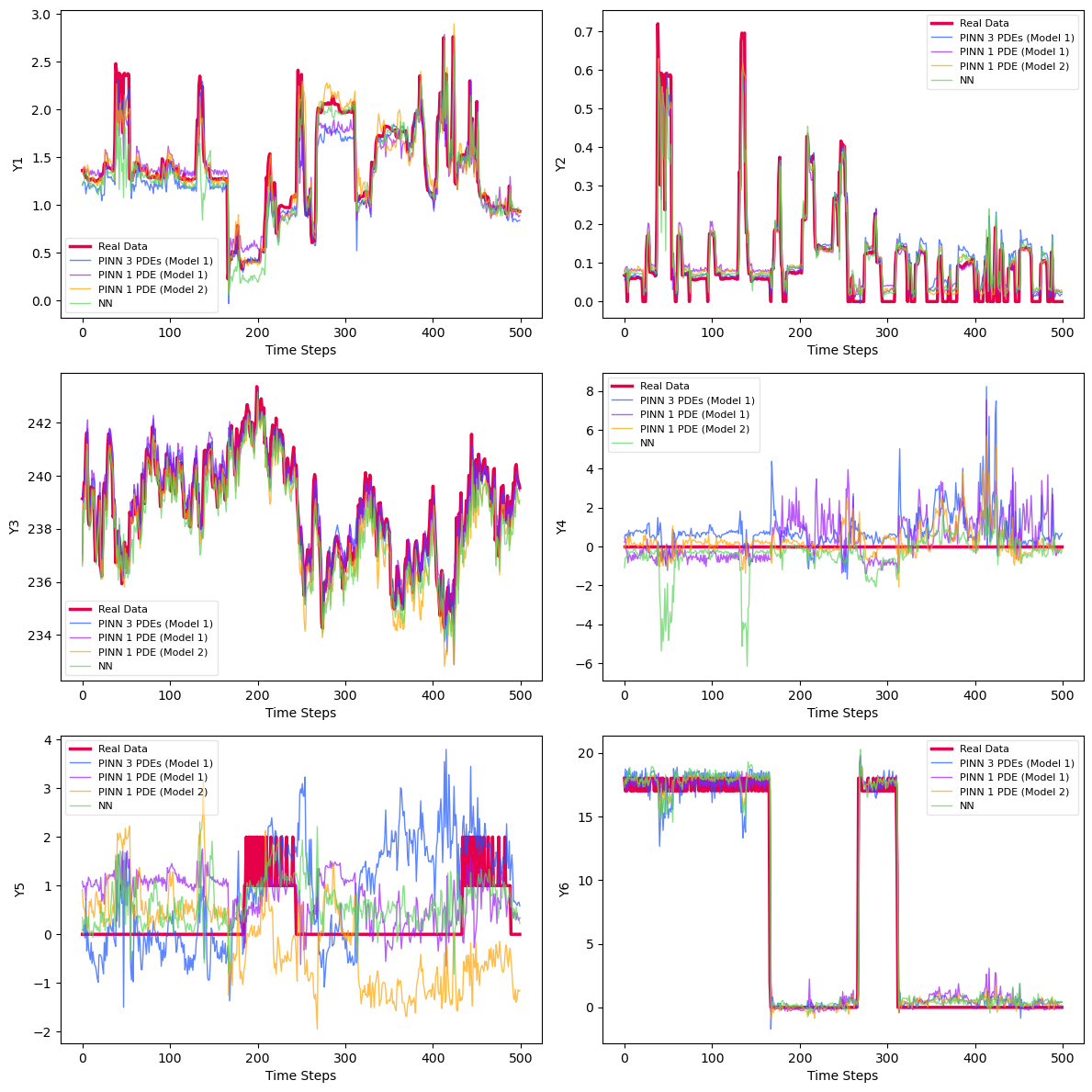} 
        \caption{Predictions of the medium-size samples with Early Stopping.}
        \label{PINN_EarlyStop_26000}
    \end{subfigure}
    \vspace{0.5em} 
    \begin{subfigure}[b]{0.95\textwidth} 
        \centering
        \includegraphics[width=\textwidth, height=9cm]{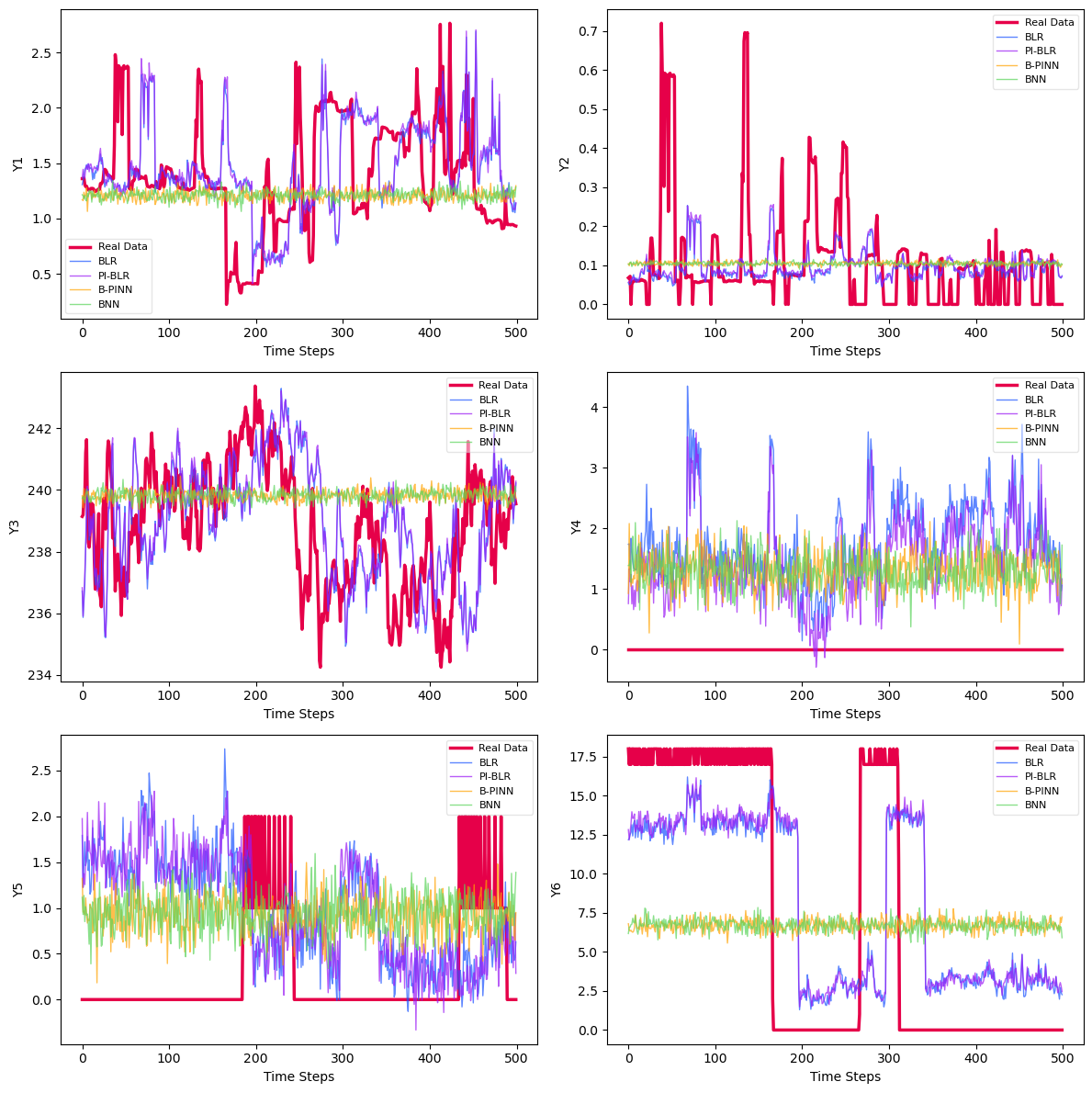} 
        \caption{Bayesian predictions of the medium-size samples.}
        \label{Bayesian_26000}
    \end{subfigure}
    \caption{Continues.} 
    \label{Plotsss}
\end{figure}
\begin{figure}[h]
    \ContinuedFloat
    \centering
    \begin{subfigure}[b]{0.95\textwidth} 
        \centering
        \includegraphics[width=\textwidth, height=9cm]{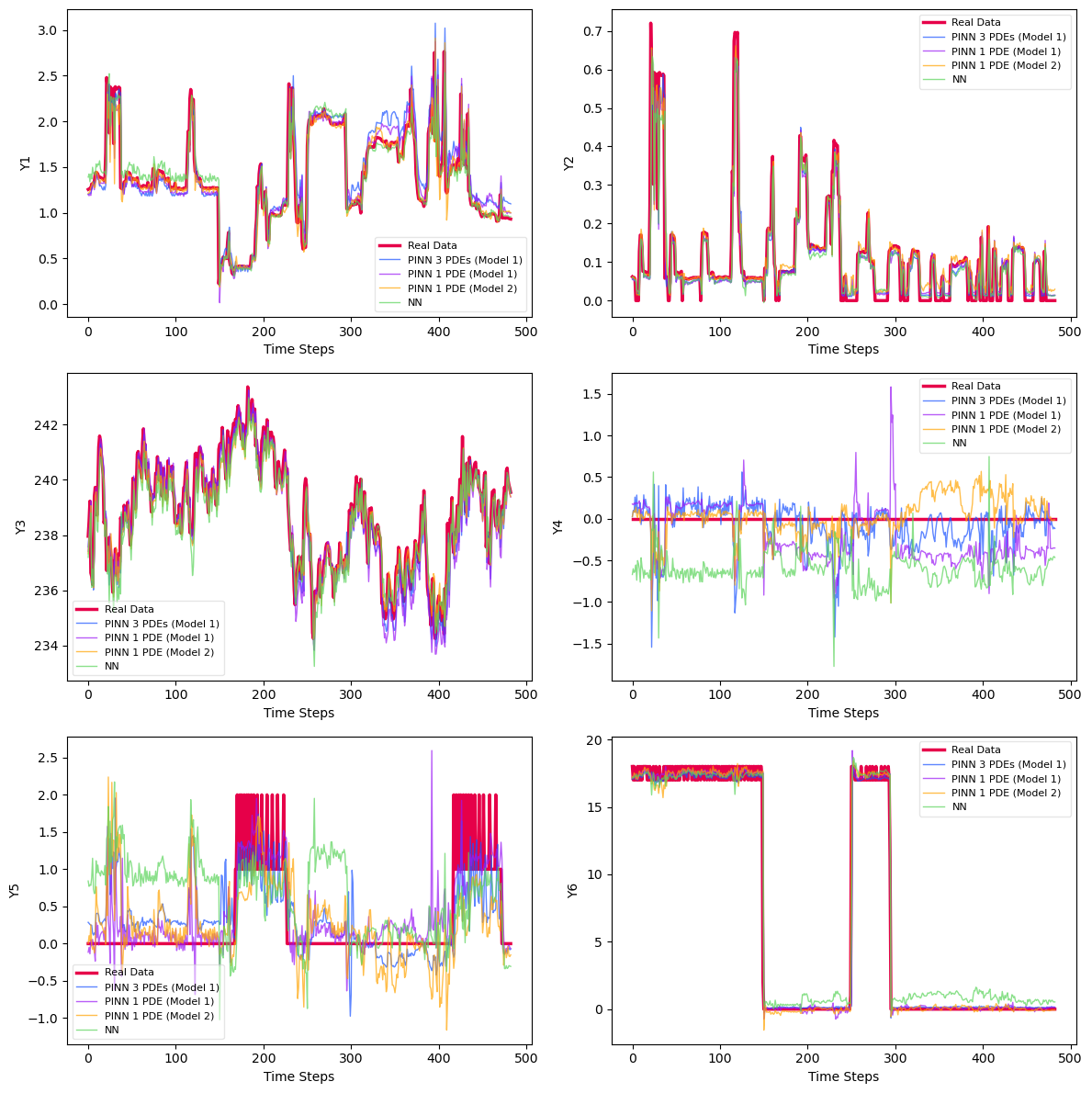} 
        \caption{Predictions of the large samples with $50$ epochs.}
        \label{PINN_50Epochs_91656}
    \end{subfigure}
    \vspace{0.5em} 
    \begin{subfigure}[b]{0.95\textwidth} 
        \centering
        \includegraphics[width=\textwidth, height=9cm]{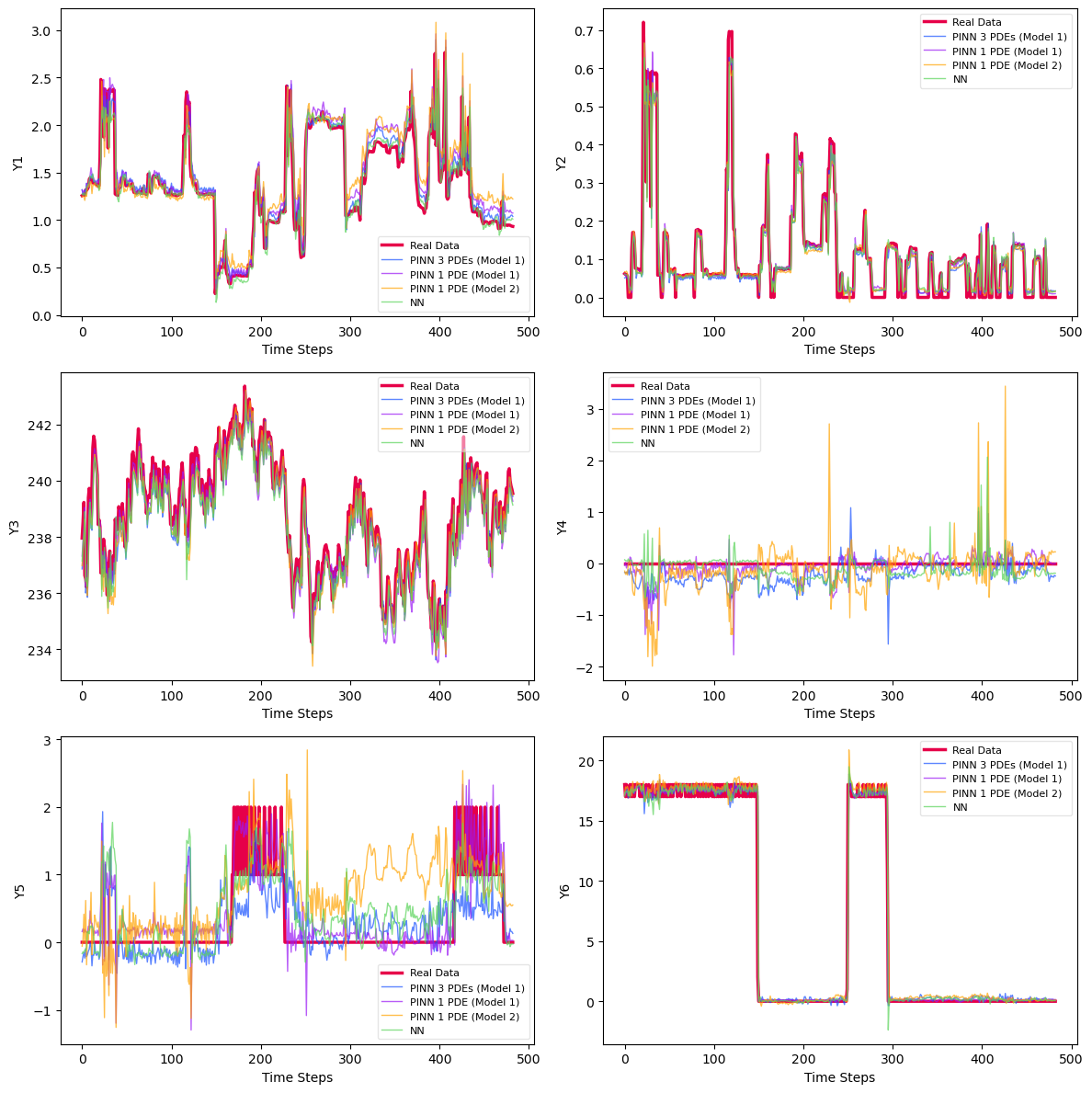} 
        \caption{Predictions of the large samples with early stopping.}
        \label{PINN_EarlyStop_91656}
    \end{subfigure}
    \caption{Continues.} 
    \label{Plotsss}
\end{figure}
\clearpage
\begin{figure}[h]
    \ContinuedFloat
    \centering
    \begin{subfigure}[b]{0.95\textwidth} 
        \centering
        \includegraphics[width=\textwidth, height=9cm]{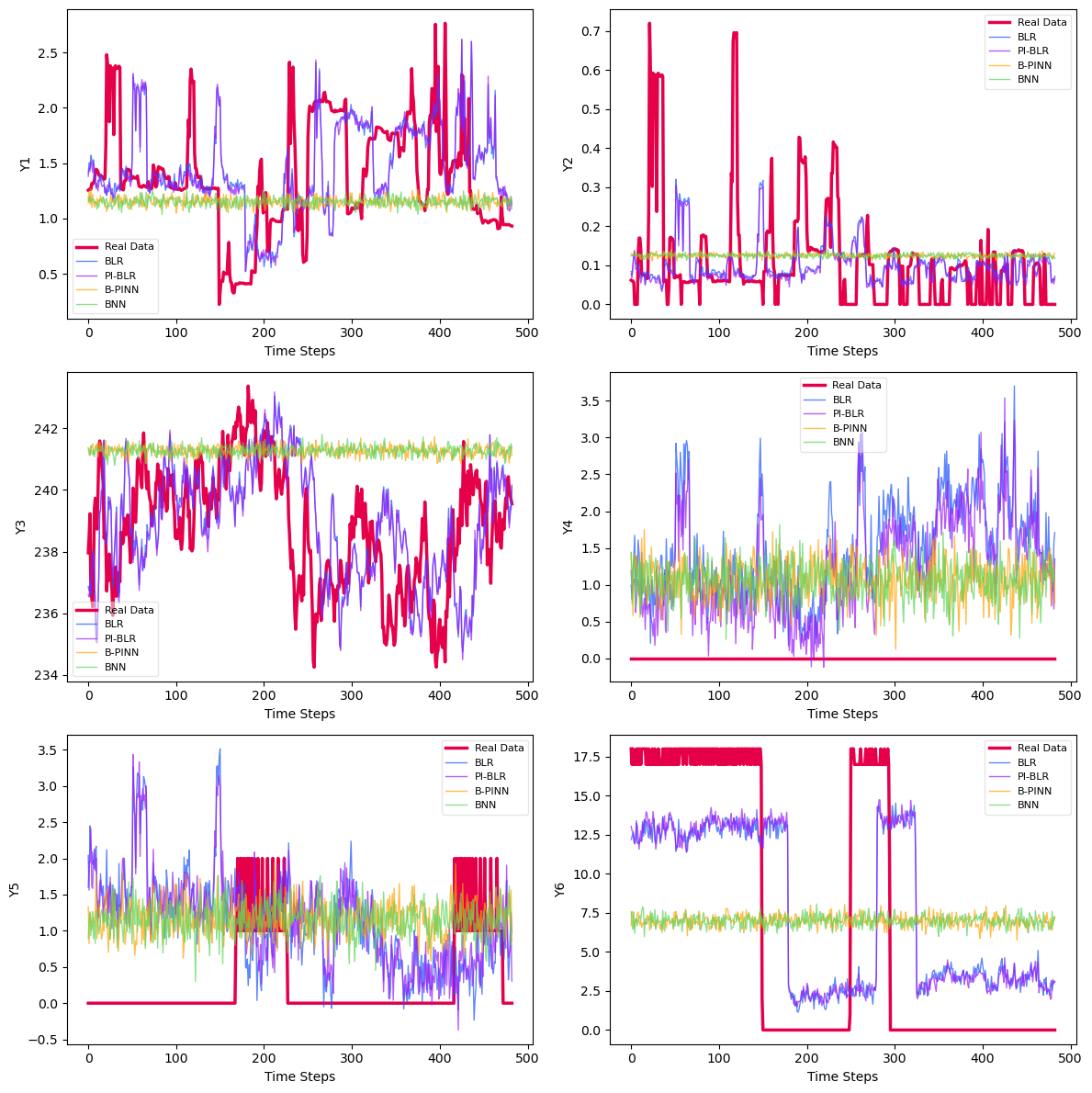} 
        \caption{Bayesian predictions of the large samples.}
        \label{Bayesian_91656}
    \end{subfigure}
    \caption{Prediction Comparison methods across different sample sizes with 50 epochs, early stopping, and Bayesian predictions, showcasing the model performances for small, medium, and large samples.}
    \label{Plotsss}
\end{figure}

Since TCNs have a strong architecture for recognizing temporal relationships, we also apply a TCN in the prediction phase.
The TCN layer extracts temporal features through dilated convolutions with dilation rates of $[1, 2, 4, 8]$, utilizing $64$ filters and a kernel size of $2$. The output layer then maps these features to $6$ output variables using a dense layer.
We use the same architecture for both PINNs and the standard NNs. The key differences in the PINNs arise from the selection of various PDEs. While the training for PINNs focuses on minimizing \eqref{PINN_Loss}, the NNs primarily aim to reduce data loss.

To facilitate comparison across sample sizes in the prediction phase, we selected three different sample sizes for the training, validation, and testing procedures. The starting point for the training set is the next data point following the last sample of the PDE extraction phase, resulting in distinct sample values. We define small ($2000$ training, $500$ validation, and $500$ testing), medium-size ($26000$ training, $4000$ validation, and $500$ testing), and large ($91656$ training, $4324$ validation, and $500$ testing) sample sizes. The large samples consist of the remaining data set after the PDE extraction phase.

The final prediction results are presented in Table \ref{PredictionPhaseMetrics} and Figure \ref{Plotsss}. Generally, the non-Bayesian methods perform better and are approximately faster by over $98\%$. One reason for this is that Bayesian models are more complex and require significantly more samples and chains to train their networks accurately. The poor results are why we do not train the Bayesian network using the PDEs from model $2$. Among the Bayesian methods, the highest $R^2$ is achieved by the B-PINN, which utilizes large sample sizes, while the best plot is produced by the BLR and PI-BLR, as shown in Figure \ref{Bayesian_91656}.

We train the non-Bayesian networks for $50$ epochs, and to facilitate a better comparison, we include results for early stopping with a maximum of $200$ epochs; however, all training terminates in fewer than $50$ epochs. In all cases, the methods struggle with the predictions of $Y_4$, resulting in an $R^2$ of $0$.

For all other variables, excluding $Y_1$, the PINNs yield the best predictions, with the highest $R^2$ values highlighted in bold in Table \ref{PredictionPhaseMetrics}. The best prediction method for $Y_1$ is the simple NN.

PINNs are well-known for providing additional information to the training process, enhancing efficiency. In this research, we do not have predefined PDEs; instead, we extract them from historical data. By incorporating these PDEs, the efficiency of the network improves. In other words, the external computations during the PDE extraction phase add valuable information to the network, thereby enhancing its efficiency, even though this information is derived from the available data.

\section{Conclusion}\label{S5}
This work demonstrates that physics-informed learning frameworks can be effectively adapted to systems with unknown governing equations by integrating data-derived PDEs. We show that surrogate TCNs with dilated convolutions can automatically extract meaningful PDEs from raw MTS data. This capability enables the application of PINNs even in domains where predefined physical laws are absent.
While PINNs achieved significant improvements in predictive accuracy for most variables, Bayesian methods revealed trade-offs between computational efficiency and uncertainty quantification. However, Bayesian methods required rich datasets along with sufficient sampling size and chains. Notably, the extracted PDEs improved model efficiency by steering predictions toward physically plausible solutions, even when derived solely from historical data. These findings highlight the potential of hybrid data-physics approaches in fields such as economics and social sciences, where explicit governing equations are often difficult to identify.

\section*{\bf Declarations}
This work was supported by the Innovative Research Group Project of the National Natural Science Foundation of China (Grant No. 51975253). The authors report no conflicts of interest or personal ties that may affect the work in this paper. The contributions are as follows:\\
Seyedeh Azadeh Fallah Mortezanejad: Methodology, Conceptualization, Validation, Writing-original draft, Writing-review and editing.\\
Ruochen Wang: Supervision, Methodology, Investigation, Writing-original
draft, Writing-review and editing.\\
Ali Mohammad-Djafari: Supervision, Methodology, Writing-review and editing.

\bibliographystyle{amsplain}

\begin{thebibliography}{5}
\bibitem[{Bararnia and Esmaeilpour(2022)}]{bararnia2022application}
\bibinfo{author}{Bararnia, H.}, \bibinfo{author}{Esmaeilpour, M.},
  \bibinfo{year}{2022}.
\newblock \bibinfo{title}{On the application of physics informed neural
  networks (pinn) to solve boundary layer thermal-fluid problems}.
\newblock \bibinfo{journal}{International Communications in Heat and Mass
  Transfer} \bibinfo{volume}{132}, \bibinfo{pages}{105890}.
\bibitem[{Brunton et~al.(2016)Brunton, Proctor and
  Kutz}]{brunton2016discovering}
\bibinfo{author}{Brunton, S.L.}, \bibinfo{author}{Proctor, J.L.},
  \bibinfo{author}{Kutz, J.N.}, \bibinfo{year}{2016}.
\newblock \bibinfo{title}{Discovering governing equations from data by sparse
  identification of nonlinear dynamical systems}.
\newblock \bibinfo{journal}{Proceedings of the national academy of sciences}
  \bibinfo{volume}{113}, \bibinfo{pages}{3932--3937}.
\bibitem[{Changdar et~al.(2024)Changdar, Bhaumik, Sadhukhan, Pandey,
  Mukhopadhyay, De and Bakalis}]{changdar2024integrating}
\bibinfo{author}{Changdar, S.}, \bibinfo{author}{Bhaumik, B.},
  \bibinfo{author}{Sadhukhan, N.}, \bibinfo{author}{Pandey, S.},
  \bibinfo{author}{Mukhopadhyay, S.}, \bibinfo{author}{De, S.},
  \bibinfo{author}{Bakalis, S.}, \bibinfo{year}{2024}.
\newblock \bibinfo{title}{Integrating symbolic regression with physics-informed
  neural networks for simulating nonlinear wave dynamics in arterial blood
  flow}.
\newblock \bibinfo{journal}{Physics of Fluids} \bibinfo{volume}{36}.
\bibitem[{Chen et~al.(2021)Chen, Guo, Zhang and Pan}]{chen2021partially}
\bibinfo{author}{Chen, J.}, \bibinfo{author}{Guo, Z.}, \bibinfo{author}{Zhang,
  L.}, \bibinfo{author}{Pan, J.}, \bibinfo{year}{2021}.
\newblock \bibinfo{title}{A partially confirmatory approach to scale
  development with the bayesian lasso.}
\newblock \bibinfo{journal}{Psychological Methods} \bibinfo{volume}{26},
  \bibinfo{pages}{210}.
\bibitem[{Fasel et~al.(2022)Fasel, Kutz, Brunton and
  Brunton}]{fasel2022ensemble}
\bibinfo{author}{Fasel, U.}, \bibinfo{author}{Kutz, J.N.},
  \bibinfo{author}{Brunton, B.W.}, \bibinfo{author}{Brunton, S.L.},
  \bibinfo{year}{2022}.
\newblock \bibinfo{title}{Ensemble-sindy: Robust sparse model discovery in the
  low-data, high-noise limit, with active learning and control}.
\newblock \bibinfo{journal}{Proceedings of the Royal Society A}
  \bibinfo{volume}{478}, \bibinfo{pages}{20210904}.
\bibitem[{Fraza et~al.(2021)Fraza, Dinga, Beckmann and
  Marquand}]{fraza2021warped}
\bibinfo{author}{Fraza, C.J.}, \bibinfo{author}{Dinga, R.},
  \bibinfo{author}{Beckmann, C.F.}, \bibinfo{author}{Marquand, A.F.},
  \bibinfo{year}{2021}.
\newblock \bibinfo{title}{Warped bayesian linear regression for normative
  modelling of big data}.
\newblock \bibinfo{journal}{NeuroImage} \bibinfo{volume}{245},
  \bibinfo{pages}{118715}.
\bibitem[{Gholipourshahraki et~al.(2024)Gholipourshahraki, Bai, Shrestha,
  Hjelholt, Hu, Kjolby, Rohde and
  S{\o}rensen}]{gholipourshahraki2024evaluation}
\bibinfo{author}{Gholipourshahraki, T.}, \bibinfo{author}{Bai, Z.},
  \bibinfo{author}{Shrestha, M.}, \bibinfo{author}{Hjelholt, A.},
  \bibinfo{author}{Hu, S.}, \bibinfo{author}{Kjolby, M.},
  \bibinfo{author}{Rohde, P.D.}, \bibinfo{author}{S{\o}rensen, P.},
  \bibinfo{year}{2024}.
\newblock \bibinfo{title}{Evaluation of bayesian linear regression models for
  gene set prioritization in complex diseases}.
\newblock \bibinfo{journal}{PLoS genetics} \bibinfo{volume}{20},
  \bibinfo{pages}{e1011463}.
\bibitem[{Hu et~al.(2024)Hu, Qi and Chao}]{hu2024physics}
\bibinfo{author}{Hu, H.}, \bibinfo{author}{Qi, L.}, \bibinfo{author}{Chao, X.},
  \bibinfo{year}{2024}.
\newblock \bibinfo{title}{Physics-informed neural networks (pinn) for
  computational solid mechanics: Numerical frameworks and applications}.
\newblock \bibinfo{journal}{Thin-Walled Structures} , \bibinfo{pages}{112495}.
\bibitem[{Liang et~al.(2024)Liang, Tiwari, Nowaczyk and
  Byttner}]{liang2024higher}
\bibinfo{author}{Liang, G.}, \bibinfo{author}{Tiwari, P.},
  \bibinfo{author}{Nowaczyk, S.}, \bibinfo{author}{Byttner, S.},
  \bibinfo{year}{2024}.
\newblock \bibinfo{title}{Higher-order spatio-temporal physics-incorporated
  graph neural network for multivariate time series imputation}, in:
  \bibinfo{booktitle}{Proceedings of the 33rd ACM International Conference on
  Information and Knowledge Management}, pp. \bibinfo{pages}{1356--1366}.
\bibitem[{Liu et~al.(2025)Liu, Shen, Yang and Yang}]{liu2025cnn}
\bibinfo{author}{Liu, Y.Y.}, \bibinfo{author}{Shen, J.X.},
  \bibinfo{author}{Yang, P.P.}, \bibinfo{author}{Yang, X.W.},
  \bibinfo{year}{2025}.
\newblock \bibinfo{title}{A cnn-pinn-drl driven method for shape optimization
  of airfoils}.
\newblock \bibinfo{journal}{Engineering Applications of Computational Fluid
  Mechanics} \bibinfo{volume}{19}, \bibinfo{pages}{2445144}.
\bibitem[{Ma et~al.(2024)Ma, Fu, Guo and Zhai}]{ma2024incorporating}
\bibinfo{author}{Ma, M.}, \bibinfo{author}{Fu, L.}, \bibinfo{author}{Guo, X.},
  \bibinfo{author}{Zhai, Z.}, \bibinfo{year}{2024}.
\newblock \bibinfo{title}{Incorporating lasso regression to physics-informed
  neural network for inverse pde problem.}
\newblock \bibinfo{journal}{CMES-Computer Modeling in Engineering \& Sciences}
  \bibinfo{volume}{141}.
\bibitem[{Mohammad-Djafari(2025)}]{mohammad2025bayesian}
\bibinfo{author}{Mohammad-Djafari, A.}, \bibinfo{year}{2025}.
\newblock \bibinfo{title}{Bayesian physics informed neural networks for linear
  inverse problems}.
\newblock \bibinfo{journal}{arXiv preprint arXiv:2502.13827} .
\bibitem[{Nayek et~al.(2021)Nayek, Fuentes, Worden and Cross}]{nayek2021spike}
\bibinfo{author}{Nayek, R.}, \bibinfo{author}{Fuentes, R.},
  \bibinfo{author}{Worden, K.}, \bibinfo{author}{Cross, E.J.},
  \bibinfo{year}{2021}.
\newblock \bibinfo{title}{On spike-and-slab priors for bayesian equation
  discovery of nonlinear dynamical systems via sparse linear regression}.
\newblock \bibinfo{journal}{Mechanical Systems and Signal Processing}
  \bibinfo{volume}{161}, \bibinfo{pages}{107986}.
\bibitem[{Park and Casella(2008)}]{park2008bayesian}
\bibinfo{author}{Park, T.}, \bibinfo{author}{Casella, G.},
  \bibinfo{year}{2008}.
\newblock \bibinfo{title}{The bayesian lasso}.
\newblock \bibinfo{journal}{Journal of the american statistical association}
  \bibinfo{volume}{103}, \bibinfo{pages}{681--686}.
\bibitem[{Peitz et~al.(2024)Peitz, Stenner, Chidananda, Wallscheid, Brunton and
  Taira}]{peitz2024distributed}
\bibinfo{author}{Peitz, S.}, \bibinfo{author}{Stenner, J.},
  \bibinfo{author}{Chidananda, V.}, \bibinfo{author}{Wallscheid, O.},
  \bibinfo{author}{Brunton, S.L.}, \bibinfo{author}{Taira, K.},
  \bibinfo{year}{2024}.
\newblock \bibinfo{title}{Distributed control of partial differential equations
  using convolutional reinforcement learning}.
\newblock \bibinfo{journal}{Physica D: Nonlinear Phenomena}
  \bibinfo{volume}{461}, \bibinfo{pages}{134096}.
\bibitem[{Perumal et~al.(2023)Perumal, Abueidda, Koric and
  Kontsos}]{perumal2023temporal}
\bibinfo{author}{Perumal, V.}, \bibinfo{author}{Abueidda, D.},
  \bibinfo{author}{Koric, S.}, \bibinfo{author}{Kontsos, A.},
  \bibinfo{year}{2023}.
\newblock \bibinfo{title}{Temporal convolutional networks for data-driven
  thermal modeling of directed energy deposition}.
\newblock \bibinfo{journal}{Journal of Manufacturing Processes}
  \bibinfo{volume}{85}, \bibinfo{pages}{405--416}.
\bibitem[{Raissi et~al.(2019)Raissi, Perdikaris and
  Karniadakis}]{raissi2019physics}
\bibinfo{author}{Raissi, M.}, \bibinfo{author}{Perdikaris, P.},
  \bibinfo{author}{Karniadakis, G.E.}, \bibinfo{year}{2019}.
\newblock \bibinfo{title}{Physics-informed neural networks: A deep learning
  framework for solving forward and inverse problems involving nonlinear
  partial differential equations}.
\newblock \bibinfo{journal}{Journal of Computational physics}
  \bibinfo{volume}{378}, \bibinfo{pages}{686--707}.
\bibitem[{Repository(2020)}]{electric_power_consumption}
\bibinfo{author}{Repository, U.M.L.}, \bibinfo{year}{2020}.
\newblock \bibinfo{title}{Household electric power consumption}.
\newblock
  \bibinfo{howpublished}{\url{https://www.kaggle.com/datasets/uciml/electric-power-consumption-data-set}}.
\newblock \bibinfo{note}{Accessed: 2023-03-03}.
\bibitem[{Schmid et~al.(2024)Schmid, Doostan and
  Pourahmadian}]{schmid2024ensemble}
\bibinfo{author}{Schmid, A.C.}, \bibinfo{author}{Doostan, A.},
  \bibinfo{author}{Pourahmadian, F.}, \bibinfo{year}{2024}.
\newblock \bibinfo{title}{Ensemble wsindy for data driven discovery of
  governing equations from laser-based full-field measurements}.
\newblock \bibinfo{journal}{arXiv preprint arXiv:2409.20510} .
\bibitem[{Tadmor(2012)}]{tadmor2012review}
\bibinfo{author}{Tadmor, E.}, \bibinfo{year}{2012}.
\newblock \bibinfo{title}{A review of numerical methods for nonlinear partial
  differential equations}.
\newblock \bibinfo{journal}{Bulletin of the American Mathematical Society}
  \bibinfo{volume}{49}, \bibinfo{pages}{507--554}.
\bibitem[{Tibshirani(1996)}]{tibshirani1996regression}
\bibinfo{author}{Tibshirani, R.}, \bibinfo{year}{1996}.
\newblock \bibinfo{title}{Regression shrinkage and selection via the lasso}.
\newblock \bibinfo{journal}{Journal of the Royal Statistical Society Series B:
  Statistical Methodology} \bibinfo{volume}{58}, \bibinfo{pages}{267--288}.
\bibitem[{Wang et~al.(2022)Wang, Jiang, Song, Fu, Zhang, Chen and
  Zhang}]{wang2022applications}
\bibinfo{author}{Wang, D.}, \bibinfo{author}{Jiang, X.}, \bibinfo{author}{Song,
  Y.}, \bibinfo{author}{Fu, M.}, \bibinfo{author}{Zhang, Z.},
  \bibinfo{author}{Chen, X.}, \bibinfo{author}{Zhang, M.},
  \bibinfo{year}{2022}.
\newblock \bibinfo{title}{Applications of physics-informed neural network for
  optical fiber communications}.
\newblock \bibinfo{journal}{IEEE Communications Magazine} \bibinfo{volume}{60},
  \bibinfo{pages}{32--37}.
\bibitem[{Yang et~al.(2021)Yang, Meng and Karniadakis}]{yang2021b}
\bibinfo{author}{Yang, L.}, \bibinfo{author}{Meng, X.},
  \bibinfo{author}{Karniadakis, G.E.}, \bibinfo{year}{2021}.
\newblock \bibinfo{title}{B-pinns: Bayesian physics-informed neural networks
  for forward and inverse pde problems with noisy data}.
\newblock \bibinfo{journal}{Journal of Computational Physics}
  \bibinfo{volume}{425}, \bibinfo{pages}{109913}.
\bibitem[{Zhan et~al.(2024)Zhan, Guo, Yan, Chen, Chang, Babovic and
  Zheng}]{zhan2024physics}
\bibinfo{author}{Zhan, Y.}, \bibinfo{author}{Guo, Z.}, \bibinfo{author}{Yan,
  B.}, \bibinfo{author}{Chen, K.}, \bibinfo{author}{Chang, Z.},
  \bibinfo{author}{Babovic, V.}, \bibinfo{author}{Zheng, C.},
  \bibinfo{year}{2024}.
\newblock \bibinfo{title}{Physics-informed identification of pdes with lasso
  regression, examples of groundwater-related equations}.
\newblock \bibinfo{journal}{Journal of Hydrology} , \bibinfo{pages}{131504}.
\bibitem[{Zwillinger and Dobrushkin(2021)}]{zwillinger2021handbook}
\bibinfo{author}{Zwillinger, D.}, \bibinfo{author}{Dobrushkin, V.},
  \bibinfo{year}{2021}.
\newblock \bibinfo{title}{Handbook of differential equations}.
\newblock \bibinfo{publisher}{Chapman and Hall/CRC}.
\end{thebibliography}

\end{document}